\documentclass[runningheads]{llncs}

\usepackage{eccv}

\usepackage{eccvabbrv}

\usepackage{graphicx}
\usepackage{booktabs}
\usepackage{pifont}
\usepackage[accsupp]{axessibility}  
\usepackage{wrapfig}
\usepackage{float}
\usepackage{dsfont}

\usepackage{hyperref}

\usepackage{orcidlink}

\newcommand{\katyatodo}[1]{\textcolor{black}{#1}}

\begin{document}
	
	\title{Source-Free Domain-Invariant \\ Performance Prediction} 
	
	\titlerunning{Source-Free Domain-Invariant Performance Prediction}
	
	\author{Ekaterina Khramtsova\inst{1}\orcidlink{0000-0001-7531-4491} \and
		Mahsa Baktashmotlagh\inst{1}\orcidlink{0000-0001-5255-8194} \and
		Guido Zuccon\inst{1}\orcidlink{0000-0003-0271-5563}\and\\
		Xi Wang\inst{2}\orcidlink{0009-0002-1724-7694}\and
		Mathieu Salzmann\inst{3}\orcidlink{0000-0002-8347-8637}
  }
	
	\authorrunning{E.Khramtsova et al.}
	
	\institute{
        The University of Queensland, Australia \and 
        Neusoft, China \and 
        École Polytechnique Fédérale de Lausanne (EPFL), Switzerland \\
        \email{e.khramtsova@uq.edu.au}\\
        \url{https://github.com/khramtsova/source_free_pp/}
        }

	\maketitle	
	\begin{abstract}
Accurately estimating model performance poses a significant challenge, particularly in scenarios where the source and target domains follow different data distributions. Most existing performance prediction methods heavily rely on the source data in their estimation process, limiting their applicability in a more realistic setting where only the trained model is accessible. The few methods that do not require source data exhibit considerably inferior performance.
In this work, we propose a source-free approach centred on uncertainty-based estimation, using a generative
model for calibration in the absence of source data. We establish connections between our approach for unsupervised calibration and temperature scaling.
We then employ a gradient-based strategy to evaluate the correctness of the calibrated predictions.
Our experiments on benchmark object recognition datasets reveal that existing source-based methods fall short with limited source sample availability. Furthermore, our approach significantly outperforms the current state-of-the-art source-free and source-based methods, affirming its effectiveness in domain-invariant performance estimation. 


	\end{abstract}

\begin{wrapfigure}{r}{0.42\textwidth} 
	\vspace{-7ex}
	\centering
	\includegraphics[width=0.94\linewidth]{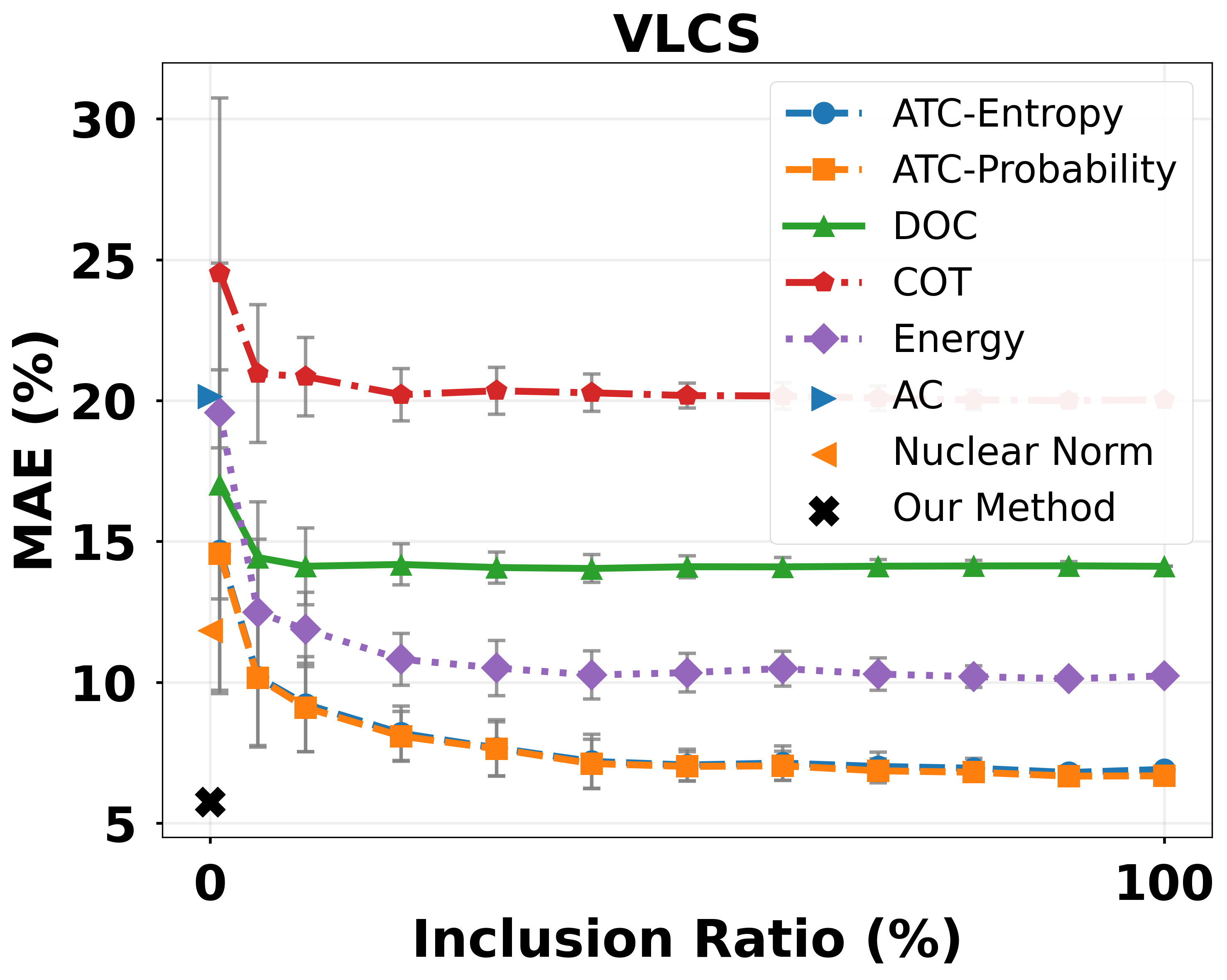} 
 	\vspace{-1ex}
	\captionof{figure}{
MAE of estimation across different percentages of 
source data. 
Our source-free method outperforms source-based baselines.
}
	\label{intro_openness}
		\vspace{-6ex}
\end{wrapfigure}

\section{Introduction}
\label{sec:intro}

Accurately estimating the performance of a model on some target (i.e., test) data is a challenging problem, particularly in the presence of a discrepancy between the target and source (i.e., training) data distributions, since it invalidates the standard cross-validation strategy. This situation, however, occurs in many practical scenarios, as manually annotating data is expensive and cannot be achieved to cover all possible variations of the input space. Predicting a model's performance on new data is thus critical to assess its reliability in the real world.


The majority of existing approaches to performance prediction ~\cite{ATC, DOC, lu2023cot_for_ood, peng2024energybased, Deng2021WhatDR, deng2020labels}  
rely on the availability of the source data as a reference point.
For instance, ATC~\cite{ATC} estimates an entropy threshold derived from the source data; COT~\cite{lu2023cot_for_ood} employs the Earth Mover's Distance between labels from the source domain and predictions from the target domain. 
However, the reliance on source data poses several challenges, especially in applications where data access is limited or the data is sensitive. For example, in critical areas such as medical diagnostics, privacy concerns limit data disclosure. In some scenarios, only a limited amount of validation data might be available. For instance, when the source dataset is small and has been used to fine-tune a network for a specific task, it is often preferable to hold out less data for validation. 

Furthermore, many performance-estimation baselines~\cite{ATC,DOC,lu2023cot_for_ood} require the presence of incorrectly-predicted validation samples within the validation set. This requirement is essential to establish a threshold to differentiate between correctly and incorrectly predicted samples. 
 Typically, the validation set acts as an indicator of convergence; thus, well-optimized networks should have a small number of incorrectly-predicted validation samples. Consequently, the effectiveness of existing methods significantly diminishes when the availability of source samples is limited. As illustrated in Fig.\ref{intro_openness}, a notable decline in the performance of all source-based baselines is observed when the availability of source samples, defined by an inclusion ratio, decreases from 100\% to 1\%, which is evidenced by a spike in the mean absolute error at a 1\% inclusion ratio.
 

In response to these challenges, the primary motivation of this paper is to overcome the constraints imposed by reliance on source data. We propose a more practical and realistic "source-free" setting in which we do not require access to the source data but only to the trained model. To this end, we introduce a method that relies exclusively on target data to predict model performance accurately in the absence of target labels. 
This approach faces the common challenge of neural networks exhibiting overconfidence in their predictions. To mitigate this, we introduce a novel calibration technique that diverges from traditional temperature scaling, prompted by the lack of source data in our setting. Our proposed unsupervised method leverages a generative model to reduce prediction certainty, and we show its connection to temperature scaling within the softmax function.
 Following our calibration, we evaluate the correctness of a prediction by measuring its similarity to a uniform distribution. Rather than a direct comparison, inspired by previous research~\cite{huang2021importance},
we rely on gradient norms derived from backpropagation of associated losses. 


	We conduct extensive experiments across a wide range of benchmark datasets, demonstrating that our method significantly surpasses existing state-of-the-art source-free approaches~\cite{AC, Deng2023ConfidenceAD} and competes favorably with source-based methods, despite not utilizing any source data. Furthermore, we investigate the volume of data required for source-based methods to match our performance and show that our method outperforms other baselines on several datasets, even when all the source data is available. This highlights the effectiveness of our approach in producing reliable source-free domain-invariant accuracy predictions.


\section{Related Work}



Most existing performance-prediction methods depend on source data to analyze the network's behavior or properties. They can be divided in two categories: sample-wise and dataset-wide methods.

Sample-wise methods associate each sample with a score that correlates with the quality of its prediction. For example, the ATC method by Garg et al.~\cite{ATC} calculates a threshold from the source data's validation set, based on entropy or prediction probability. This threshold aligns with the validation accuracy, and the model's performance on the unlabeled target dataset is estimated by the proportion of samples exceeding this threshold. The DOC~\cite{DOC} approach calculates the difference in confidences between source and target data, derived from predicted probabilities. 
The Confidence Optimal Transport (COT) method, introduced by Lu et al.~\cite{lu2023cot_for_ood} leverages optimal transport theory for performance prediction by calculating the Wasserstein distance between predicted target class probabilities and the true source label distribution. Inspired by the work of Grathwohl et al~\cite{Grathwohl2020Your}, who reinterpret a classifier as an energy-based model, Peng et al.~\cite{peng2024energybased} propose using energy instead of the softmax output as a representative statistic for the model's predictions.
The works of Baek et al.~\cite{baek2022agreementontheline} and Jiang et al.~\cite{jiang2022assessing} show that OOD accuracy can be estimated 
by examining the consensus among neural network classifiers.  Rosenfeld et al.~\cite{Rosenfeld2023AlmostPE} devise a disagreement loss; the authors provide theoretical guarantees for their method to produce an upper bound on the error under distribution shift. 

We have selected these methods as baselines for our approach, as they most closely align with our experimental setup.

 In contrast to sample-wise methods, dataset-wide methods aim to derive a measure applicable to an entire dataset rather than individual samples.
\katyatodo{For example, Deng et al. uncover correlations between classification accuracy and both rotation prediction accuracy~\cite{Deng2021WhatDR} and the Fréchet Inception distance~\cite{deng2020labels}.}
 Yu et al.~\cite{pmlr-v162-yu22i} introduces Projection Norm to predict a model's performance by pseudo-labeling test samples and comparing the parameters of a new model trained on these pseudo-labels to an in-distribution model. Subsequent work by Xie et al.~\cite{xie2024leveraging} shows that even without fine-tuning, gradients obtained after a single backpropagation step can serve as indicators of model performance. 
 \katyatodo{Another work by Xie et al.~\cite{xie2023on} uses feature dispersion to estimate the performance. }
 
 However, per-dataset methods are not scalable for scenarios with limited data availability. 
 These approaches either report only the correlation coefficients with accuracy, without providing a means to estimate the actual accuracy value \cite{pmlr-v162-yu22i, xie2024leveraging}, or they rely on generating a large collection of augmented source datasets and analyzing the model's behavior on these sets to predict its performance on the target dataset  \cite{deng2020labels, Deng2021WhatDR, peng2024energybased}. Furthermore, they yield a uniform prediction across all target samples without distinguishing between correctly and incorrectly predicted samples.  
 \katyatodo{For more details and experiments on dataset-wide methods, please refer to the Supplementary Material.}

In addition to source-based methods, a few source-free approaches have been developed.  
Average Confidence (AC), proposed by Hendrycks et al.~\cite{AC}, calculates the model's performance by determining the maximum confidence value from softmax probabilities on the target data. 
Deng et al.~\cite{Deng2023ConfidenceAD} employ the nuclear norm, arguing that it 
captures the confidence and diversity of the predictions.

We show that, despite having fewer constraints, current source-free approaches are significantly less effective than their source-dependent counterparts, even when the latter use only a small amount of source data. By contrast, our 
source-free method not only competes with, but also surpasses the performance of source-dependent approaches on some datasets, thereby setting a new benchmark for efficiency in source-free methodologies.




Our work is also related to that of Huang et al.~\cite{huang2021importance}, who conducted an analysis of the network gradients, backpropagated from the KL divergence between the softmax output and a uniform probability distribution. Their research revealed that OOD samples exhibit a larger gradient norm than in-distribution samples.
Although we draw inspiration from the work of Huang et al.~\cite{huang2021importance} to make a decision of prediction correctness based on its closeness to the uniform distribution, our study fundamentally differs from theirs in terms of environmental assumptions, as they assume that the labels of the target dataset do not overlap with those of the source data, and access to data, as they assume access to both in-distribution and OOD samples. 

\katyatodo{Finally, a Gaussian distribution of the logits \cite{kendall2017uncertainty} or the features \cite{mukhoti2023ddu, bui2024densitysoftmax} has also been used in uncertainty estimation task, e.g. for detecting samples from unknown classes or improving calibration. However, these methods do not estimate the actual performance value and require labeled source data to train the uncertainty quantifier, which is not available in our task.
}


\section{Methodology}



\subsection{Problem Definition}

Let $\mathcal{D}_s=\{(x_{s}, y_{s})\}_{i=1}^{n_s}$ denote a labeled source dataset, where $x_s \in \mathcal{X}_s $ is a source sample, and $y_s \in \mathcal{Y}_s = \{1, . . . , C\}$ is the associated label, distributed across $C$ classes. Let $\mathcal{D}_t=\{(x_{t_j})\}_{j=1}^{n_t}$ denote an unlabeled target dataset. We assume that the marginal distributions  of the source and target datasets are different, but the class distribution is the same: $p(y_s | x_s)\approx p(y_s | x_t)$, and $p(x_s) \neq p(x_t)$. 
The source dataset is used to train a classifier $\mathcal{G}_\theta: X_s \rightarrow \tilde{Y_s}$ - a function parameterized by $\theta$ that predicts a label for a given sample. The classification accuracy of $\mathcal{G}_\theta$ for $n$ target samples is defined as $a_t = \sum_{i\in n} (\tilde{y_i}\equiv y_i)/n$.

The classical performance prediction task \cite{deng2020labels} aims to predict the performance of the classifier $\mathcal{G}_\theta$ on the target dataset, given the labeled source data and unlabeled target data. Formally, the goal is to create an accuracy predictor $A$, such that $A (\mathcal{G}_\theta, \mathcal{D}_s, \mathcal{D}_t) \rightarrow a_t$. 

In this work, we consider a more realistic setting by adding a constraint to the classical performance estimation task. Specifically, we assume that the source data is only used for training 
and is unavailable during evaluation (hence the name "source-free"). Formally, the task of source-free performance prediction is defined as creating an accuracy predictor $A$, such that $A (\mathcal{G}_\theta, \mathcal{D}_t) \rightarrow a_t$. 

\subsection{Proposed Approach}

In the absence of target labels, we propose to assess the correctness of predictions by evaluating their uncertainty. This involves determining whether the prediction for a sample is closer to a random prediction than to the one-hot class encoding.

While seemingly simple, this, however, suffers from the fact that, in practice, neural networks often exhibit a tendency towards overconfidence, resulting in overly certain estimates. To address this problem, several methods have been proposed. One such method is calibration via temperature scaling, as described in \cite{10.5555/3305381.3305518}, where the network's certainty is reduced by dividing the logits $z$ by a fixed temperature parameter $T$, which is typically learned from the validation set of the source data.
Due to the unavailability of the source data in our setting, we propose to use an alternative approach to calibrate predictions.

Specifically, we start by reducing prediction certainty through adjusting the final probabilities according to the relative distance of the prediction to each class. To achieve this, we construct a probabilistic generative model that yields more accurate estimates of the likelihood of a target sample belonging to each class. It does so by leveraging statistics derived from the observed target distribution, as opposed to the original source distribution.
We further show that it can be regarded as an unsupervised calibration method, and establish its connection to temperature scaling.

After calibrating the predictions,
we evaluate their correctness by measuring their similarity to the uniform distribution. Instead of performing a direct comparison, we draw inspiration from \cite{huang2021importance}, which showed that gradient norms have higher informativeness than the loss values themselves, and compare gradient norms derived from the backpropagation of associated losses.   


\subsubsection{Probabilistic Generative Model.}

The main idea behind our approach is to consider the classification task as a generative model that, given the features, uses posterior probabilities to make optimal class assignments for each sample. 
To account for the domain shift between source and target, we first model the class-conditional densities $p(x_t|c_i)$ and the class priors $p(c_i)$ from the observed target distribution.  We then leverage these to calculate the posterior probabilities $p(c_i|x_t)$ using Bayes' theorem. 
Let us first define the softmax function from the perspective of a generative model.

Let $x_s$ be the feature vector associated with a source sample, and $f(x_s) = z_s = Wx_s + b$ be a linear function of $x_s$ that produces logits $z_s$.  

We can define a probabilistic generative model where the posterior probability of a sample to belong to a class $c_i$ is calculated as 

$$p(c_i|x_s) = \frac{p(x_s|c_i)p(c_i)}{p(x_s)} = \frac{p(x_s|c_i)p(c_i)}{\sum_{j \in C} p(x_s|c_j) p(c_j)} =  \frac{e^{f_i(x_s)}}{\sum_{j \in C} e^{f_j(x_s)}}\;.$$

Here, $f_i(x_s) = \ln p(x_s|c_i)p(c_i)$. 

The primary limitation of the so-called softmax function is its inability to account for distribution shifts, specifically, the discrepancy in feature distribution generated by the model for the source and target datasets, i.e., $p(x_s)\neq p(x_t)$. Therefore, we aim to replace the softmax function to reflect the  similarity (or the difference) in representations between the source and target datasets.

Assuming that the class-conditional densities follow a Gaussian distribution, the density for class $i$ is defined as
$$p(x|c_i) = \mathcal{N}(x, \mu_{z_i}, \Sigma) =\frac{\exp(- \frac{1}{2} (z - \mu_i)^T \Sigma^{-1} (z - \mu_i))}{\sqrt{(2 \pi )^C |\Sigma|}}\;,$$
with a log likelihood
\begin{equation}
		\log \mathcal{N}(x, \mu_{z_i}, \Sigma)= -\frac{1}{2} [ \ln(|\Sigma|) +  C \ln(2 \pi ) \\ + (z - \mu_i)^T \Sigma^{-1}  (z-\mu_i) ]\;.
	\label{log_fn}
\end{equation}

Then, the posterior probability becomes

\begin{equation}
p(c_i|x) = \frac{ p(c_i) \mathcal{N}(x, \mu_{z_i}, \Sigma)}{ \sum_{j \in C}p(c_j) \mathcal{N}(x, \mu_{z_j}, \Sigma) }\;.
\label{log_posterior_source}
\vspace{1ex}
\end{equation}

The calculation of posterior probability in Equation~\ref{log_posterior_source} can be adapted to account for the feature distribution shift. This adaptation involves estimating the parameters of the generative model, which include Gaussian means $\mu_{z_j}$, prior probabilities for each class $p(c_j)$, and the covariance matrix $\Sigma$, from the observed target distribution. However, accurately representing the target distribution poses a challenge due to the unavailability of the ground-truth label distribution in our setup. 


To mitigate this, we propose to group the target samples into clusters based on their pseudo labels, and calculate the corresponding mean for each cluster, resulting in mean vectors [$\mu_{z_i}^t, \ldots, \mu_{z_k}^t$].
Note that, when the number of clusters ($k$) is less than the total number of classes ($C$), it indicates the presence of classes in the target data with no samples, according to the pseudo labels. For these unrepresented classes, we assign a mean value of 0. The prior probability, which represents the importance of each Gaussian, is defined by the probability of that Gaussian itself generated by other Gaussians, i.e., $p(c_i^t)=\frac{1}{\sum_{j \in C}p(c_i^t|j)}$.  The  covariance matrix $\Sigma^t$ is calculated using  all target logits.  

Importantly, the decision boundaries resulting from our approach are inherently linear in the input space. This linearity is a direct consequence of the model assuming shared covariances among the different classes. We opt for a linear model as the work of Gaston~et al.~\cite{gaston2012active} has highlighted the risk of singularities in Gaussian Mixture Models, where one covariance matrix might shrink to zero when a few data points cluster very close to the center of a Gaussian distribution. Therefore, inspired by \cite{gaston2012active}, we use a single (equal) covariance matrix across all Gaussian components, computed on the entire dataset.

\begin{figure*}[ht]
	\centering
 
	\begin{subfigure}{0.32\textwidth}
		\includegraphics[width=\linewidth]{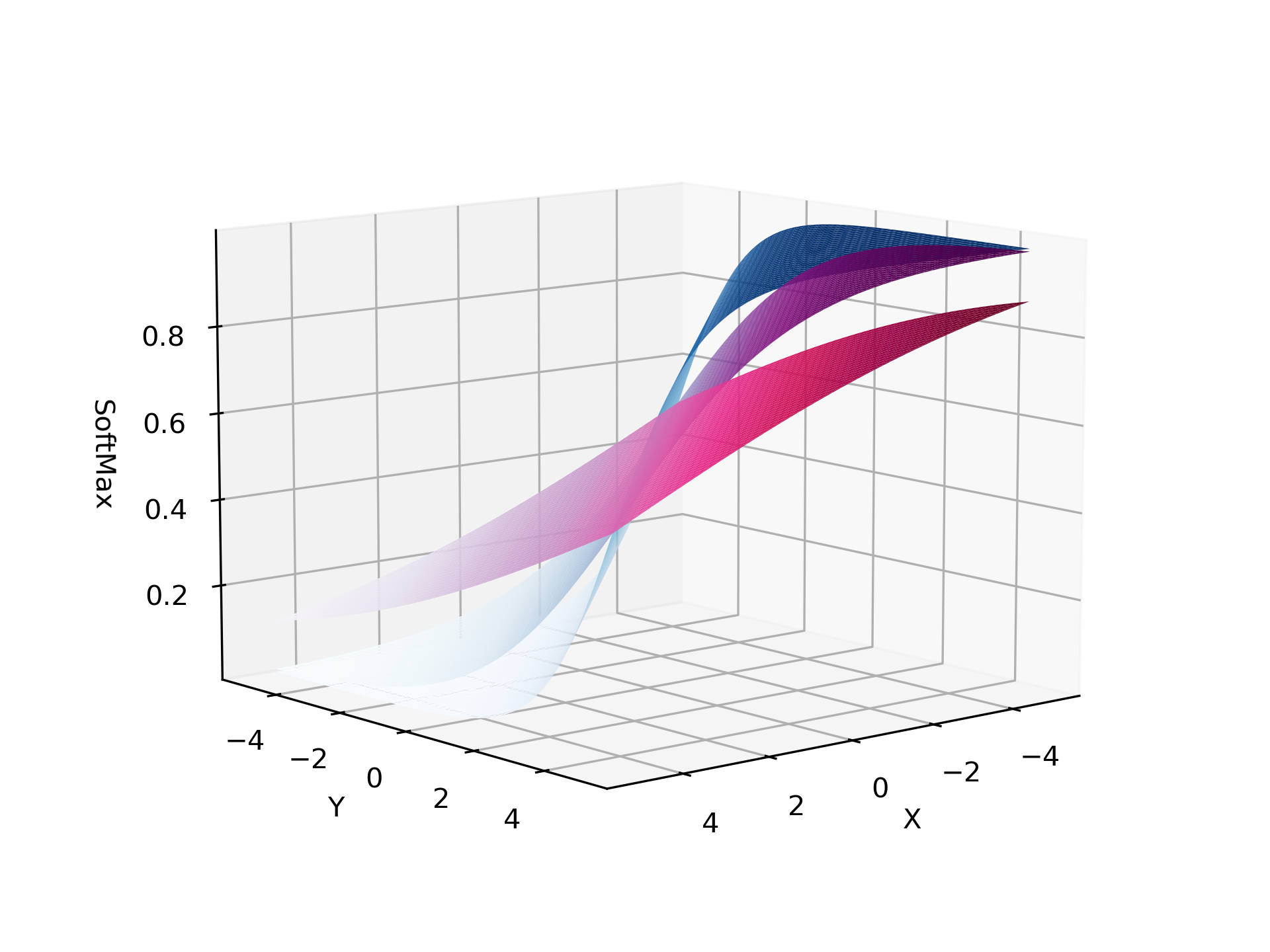}
		\caption{Softmax with Temperature}
		\label{img_smax_tpr}
	\end{subfigure}
	\begin{subfigure}{0.32\textwidth}
		\includegraphics[width=\linewidth]{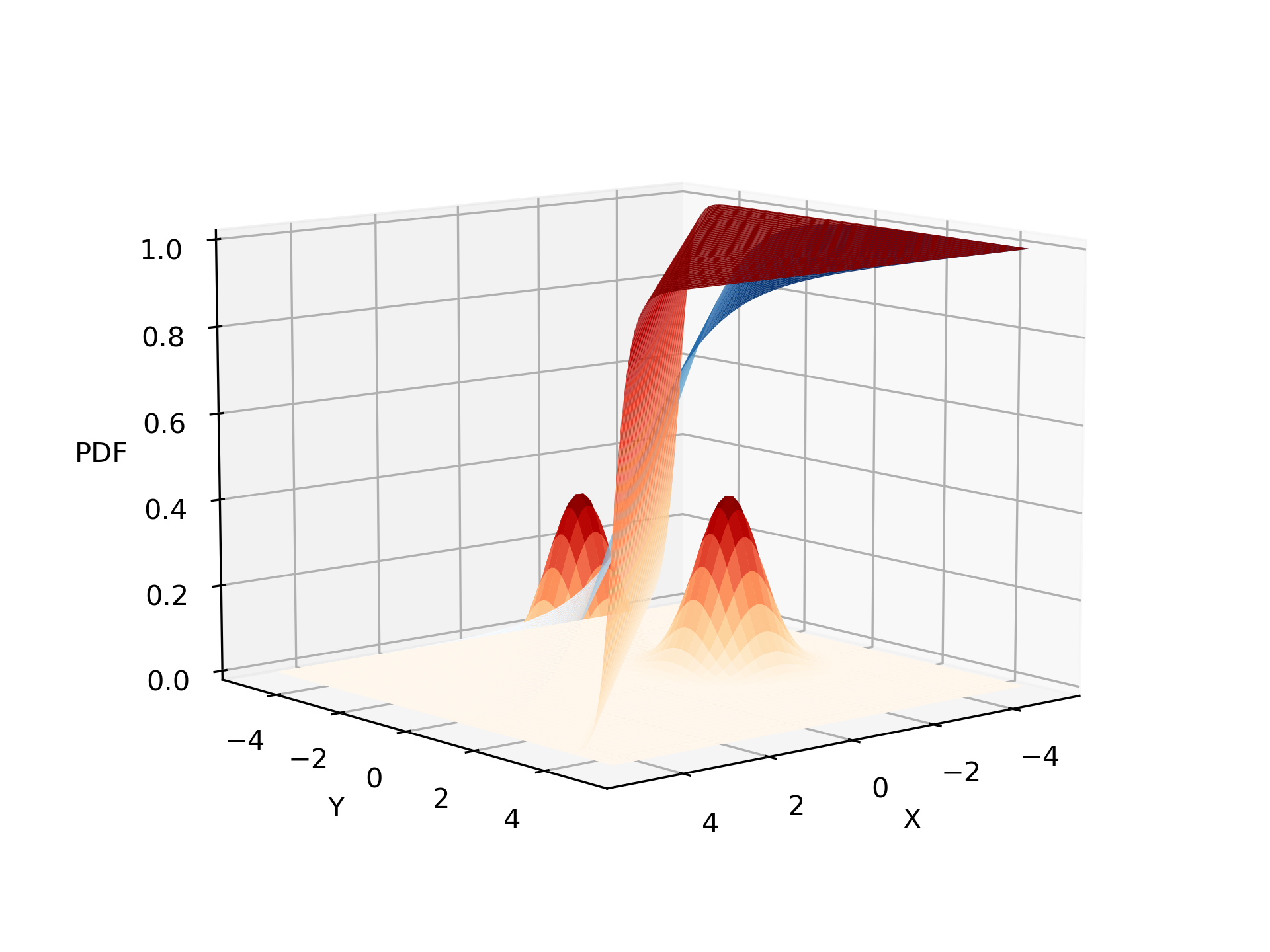}
		\caption{Generative Model}
		\label{img_mnd_1}
	\end{subfigure}
	\begin{subfigure}{0.32\textwidth}
		\includegraphics[width=\linewidth]{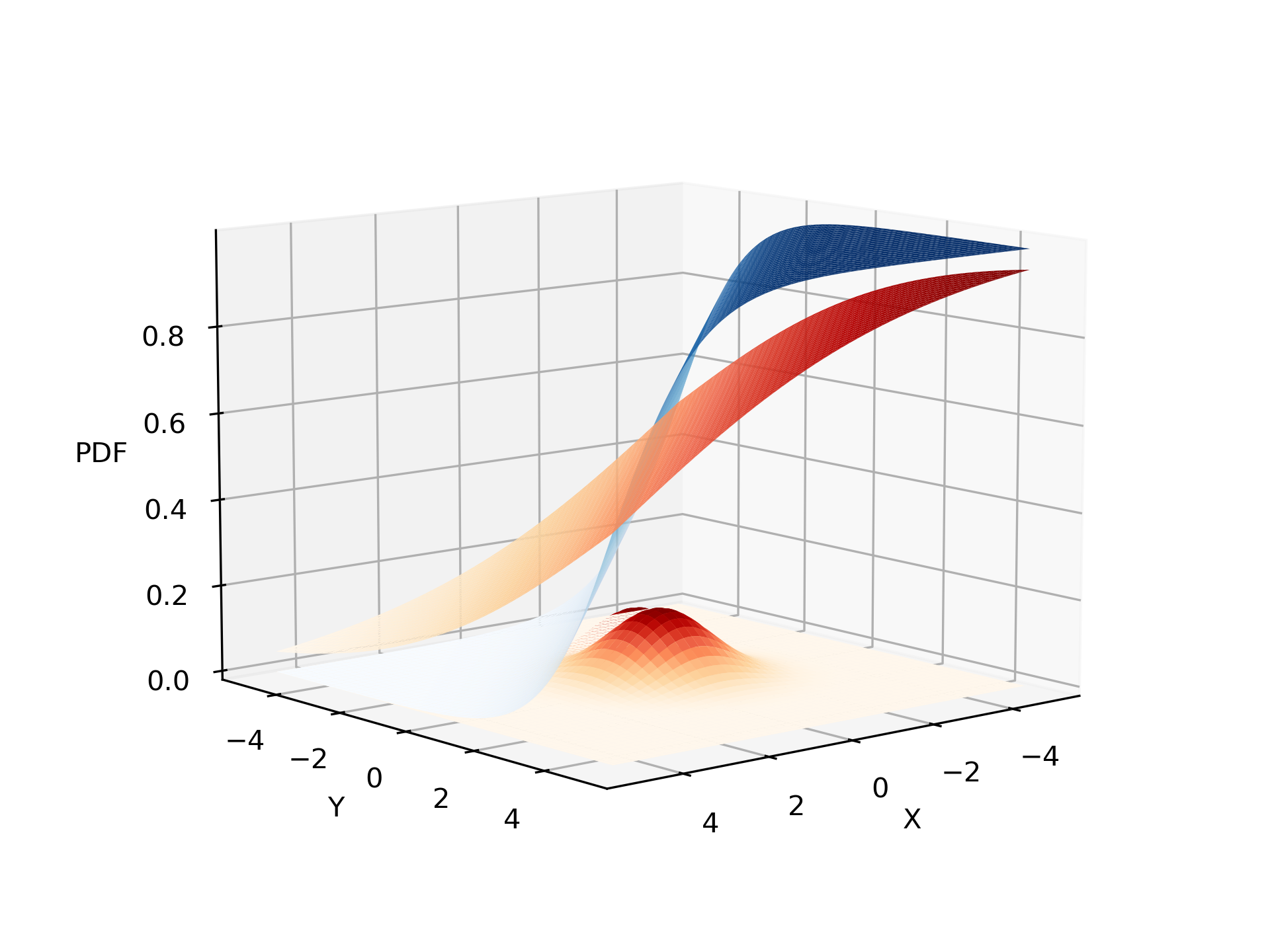}
		\caption{Generative Model}
		\label{img_mnd_2}
	\end{subfigure}
	\caption{Blue curve: Sotmax function; Orange curve: Generative Model.  }
	\vspace{-2ex}
\end{figure*}

\subsubsection{Overall Objective.} We calculate the posterior probability of the target sample by substituting the Gaussian parameters in Equation~\ref{log_posterior_source} with the parameters estimated from the observed target distribution, as discussed above. This approach yields calibrated predictions, where the calibration parameters are adjusted for each target dataset. The log likelihood of the target sample is thus computed as
\begin{equation}
	\begin{split}
		\log p(c_i|x_t) = \log \mathcal{N}(x_t, \mu_{z_i}^t, \Sigma^t)  - \log  \sum_{j  \in C, i\neq j }\mathcal{N}(\mu_{z_i}^t, \mu_{z_j}^t, \Sigma )  \\  
  -\log  \sum_{j \in C}  \sum_{i \in C, i\neq j} \mathcal{N}(x_t, \mu_{z_{j}}^t, \Sigma^t)  \mathcal{N}(\mu_{z_{j}}^t, \mu_{z_{i}}^t, \Sigma^t) \;.
	\end{split}
	\label{log_Fn}
\end{equation}

The resulting log likelihood can be directly integrated into the standard cross-entropy loss, forming our final objective function
\begin{equation}
L_{CE} (x_t) = - \sum_{i \in C} \mathds{1} (z = c_i) \log p(c_i|x_t)\;,
\end{equation}
where $ \mathds{1}$ is the indicator function, being 1 if the argument is true and 0 otherwise.

\vspace{2ex}
\noindent \textbf{Connection to Temperature Scaling.} Here, we explain the relationship between temperature scaling and our proposed generative model. According to Hinton et al.~\cite{Hinton2015DistillingTK}, one way to reduce the overconfidence of predictions is to introduce a temperature parameter to the logits. This yields a probability for class $i$ given by $p(c_i|x) = \frac{\exp(z/T)}{\sum\exp(z/T)}\;.$

In Figure \ref{img_smax_tpr}, we visualize the impact of temperature on the softmax function, by considering only two variables, X and Y, and plotting the softmax values for different temperatures. As shown by the figure, increasing the temperature results in a flatter and more uniform probability distribution.

Similarly to the temperature in the softmax, the Gaussian parameters influence the slope of the probability density function. Well-separated Gaussians with small variances and distinct means lead to sharp slopes in the density function and, consequently, more confident predictions, as shown in Figure \ref{img_mnd_1}. Conversely, similar Gaussians with close means or large variances produce flatter solutions, as depicted in Fig.~\ref{img_mnd_2}.

\vspace{2ex}
\noindent\textbf{Addressing Numerical Instability.}
When the number of classes in a dataset is large, the calculations become unstable and lead to exploding gradients. This instability is primarily caused by the sparsity of the covariance matrix $\Sigma$ as it becomes nearly singular when number of classes increases. 
Consequently, the determinant of $\Sigma$ approaches zero, leading to instability when calculating $|\Sigma^{-1}|$ in Eq.~\ref{log_fn}. 

To mitigate this, we normalize $|\Sigma^{-1}|$ using the $L_2$ norm when we have a large number of classes. Specifically, we stabilize the calculation of $ \log \sum_{j \in C} f(z, \mu_j, \Sigma)$ in Eq.~\ref{log_Fn} by writing
  $$  \log \sum_{j \in C} \mathcal{N}(x, \mu_{z_j}, \Sigma) = - \log \sqrt{2 \pi^C |\Sigma| } \\ 
    + \log \sum  \exp (- \frac{1}{2} (z -   \mu_i)  ^T\Sigma^{-1}(z - \mu_i))  \;. $$

The term $\log \sqrt{2 \pi^C |\Sigma|}$ is constant with respect to $z$ and, therefore, can be safely omitted as it does not impact the gradients. 
Note that the remaining value $ (z - \mu_i)^T\Sigma^{-1}(z - \mu_i)$ represents the squared Mahalanobis distance,  denoted as  $M^2$. To prevent potential issues related to overflow during exponentiation, we adopt the LogSumExp trick used in softmax, which yields
$$    \log \sum_{j \in C} \mathcal{N}(z, \mu_{z_j}, \Sigma) = \max (M^2)  \\
   + \log\sum   \exp (-  \frac{1}{2}   (  M^2  - \max(M^2))) \;. $$

\subsubsection{Gradient-based Performance Prediction.}

In the previous section, we adjusted the network's outputs by using the proposed probabilistic generative model. Here, we present our methodology to utilize those calibrated outputs to predict the performance on the target data. Our approach is inspired by \cite{10.5555/3305381.3305518}, which employs the information from the gradient of the loss function to estimate prediction uncertainty. Note that our aim differs from that of~\cite{10.5555/3305381.3305518}, as we seek to determine whether a sample is closer to the uniform distribution or to the one-hot class label in order to derive the correctness of a prediction. 

To achieve this, we compute two cross-entropy functions: $L_{CE}(s, PL)$ and $L_{CE}(s, U)$. 
 Here $PL$ represents the pseudo label, defined as $PL = [0, ..., 1_k,..., 0],$ $k = argmax(s) $, u is a uniform distribution $u = [1/C, 1/C, ... , 1/C ] \in \mathbb{R}^C$, and $C$ is the number of classes.

To decide whether a sample has been predicted correctly, we compare the norms of the gradients with respect to the weights of the last linear layer, denoted as $W$. Specifically, if the gradient norm is smaller for the loss based on the uniform distribution (i.e., $L_{CE}(s, U)$), we consider the sample as predicted incorrectly. Conversely, if the gradient norm is smaller for the class-based loss (i.e., $L_{CE}(s, PL)$), the sample is considered as predicted correctly.

Finally, the performance for target dataset $\mathcal{D}_t=\{(x_{t_j})\}_{j=1}^{n_t}$ is computed as 
\begin{equation}
A (\mathcal{D}_t, \mathcal{G_\theta}) = \sum_{i \in n_t}
\frac {||\frac{\partial L_{CE}(s, U)}{\partial W}||<|| \frac{\partial L_{CE}(s, PL) }{\partial W} ||}{n_t} \;.
\label{grad_norm}
\end{equation}

\section{Experiments}

\subsection{Experimental Setup}


We categorize setups into two groups based on the number of source datasets associated with each setup: single-source and multi-source. For the single-source setup, a single model is trained on the source data and then evaluated on one or multiple test sets. In the multi-source setups, each setup consists of multiple domains. We follow the leave-one-domain-out evaluation strategy: Each domain is used for training, while the remaining domains serve as test sets. 


\noindent \textbf{Single-Source Datasets:} 

\noindent\textit{Digits} consists of one source domain, MNIST~\cite{lecun2010mnist}, 
with 60K training and 10K test images of handwritten digits distributed between 10 classes, and three target datasets:  
USPS~\cite{USPS}, SVHN~\cite{SVHN}, and SYNTH~\cite{SYNTH}. The target datasets also consist of digit images, but they differ in terms of colors, styles, and backgrounds. USPS and SVHN represent natural shifts, while SYNTH represents a synthetic shift. 

 For a more extensive evaluation, we incorporate real-world datasets from the WILDs benchmark~\cite{pmlr-v139-koh21a}, namely fMoW\cite{8578744} and Camelyon17~\cite{Bndi2019FromDO}. 
fMoW includes satellite images categorized into 62 building or land classes. Camelyon17, a medical dataset, includes patches from whole-slide images of lymph node sections, divided into 2 classes: benign patches and patches with metastatic breast cancer.

\noindent\textbf{Multi-Source Datasets:}

\noindent As multi-domain datasets, we selected the following 4 datasets from Domain Bed~\cite{gulrajani2021in}: PACS~\cite{PACS}, VLCS~\cite{VLCS}, Office Home~\cite{OFFICE_HOME}, and Terra Incognita~\cite{Terra_incognita}. Terra Incognita features wild animal photographs captured across different locations, while the other datasets include images of generic categories, including people and various objects. Each dataset consists of four diverse domains that share the same class labels but significantly differ in style. 


\noindent\textbf{Networks.}
We utilize the LeNet architecture~\cite{lenet} for the Digits setup.
For both the fMoW and Camelyon17 datasets, we adopt the Densenet-121~\cite{densenet} architecture and follow the ERM training procedure from the WILDs benchmark~\cite{pmlr-v139-koh21a}. For Pacs,  VLCS, Office Home, and Terra Incognita, we use ResNet50~\cite{resnet} and follow the default training procedure from DomainBed benchmark~\cite{gulrajani2021in}.



\noindent\textbf{Methods and Evaluation Metrics.} We conduct a comprehensive comparison between our method and the following state-of-the-art (SOTA) performance estimation techniques. 
In our source-based baselines, we include ATC~\cite{ATC}, DOC~\cite{DOC}, COT~\cite{lu2023cot_for_ood}, Energy ~\cite{peng2024energybased}, and  Agree Score~\cite{baek2022agreementontheline}. For ATC, 
we use both probability and entropy as confidence scores. For source-free baselines, we evaluate AC~\cite{AC} and Nuclear Norm~\cite{Deng2023ConfidenceAD}. Note that we also consider the calibrated variant of Nuclear Norm, which significantly enhances performance but requires source data for calibration. 
Finally, we show the significance of our generative model by reporting the performance of GradNorm without the proposed calibration.

\begin{table*}[ht]
	\begin{center}
		\caption{
			Absolute Error between predicted and Ground Truth accuracy for \textbf{Single-Source datasets}. Results for source-based baselines report the mean and standard deviation across 20 trials, with each trial randomly selecting 1\% of source samples.}
		\vspace{-3ex}
		
		\label{table:single_source}
		
		\scalebox{0.85}{
			{\renewcommand{\arraystretch}{1.2}%
				\begin{tabular}{@{}l|c|ccc|c|cc|c@{}} 
					
					\toprule
					& &\multicolumn{4}{c}{Digits} & \multicolumn{3}{|c}{Wilds}\\ \midrule
					&Source Free & SVHN & USPS & SYNTH & MAE & FMoW & Camelyon& MAE \\
					\midrule
					
					
					ATC$_{e}$ &  \ding{55}  & 38.25 $_{\pm1.76}$  &	20.54 $_{\pm18.85}$ &	23.38 $_{\pm9.67}$ &  27.39   & 
					3.88 $_{\pm3.87}$&	11.4 $_{\pm3.01}$ & 7.64 
					
					\\\hline
					ATC$_{e}$-C &\ding{55} & 38.43 $_{\pm1.83}$ &	22.63 $_{\pm21.21}$ &	24.11$_{\pm10.19}$  & 28.39 & 
					
					4.10 $_{\pm3.78}$&	11.4 $_{\pm3.01}$& 7.75
					\\ \hline
					ATC$_{p}$ & \ding{55}  & 35.34 $_{\pm4.9}$&	18.09$_{\pm18.93}$&	18.89 $_{\pm12.86}$& 24.10 & 
					
					3.85 $_{\pm2.76}$&	11.4 $_{\pm3.01}$&7.63
					\\ \hline
					ATC$_{p}$-C &\ding{55} &  35.51 $_{\pm4.99}$&	19.75 $_{\pm20.93}$&	19.48 $_{\pm13.22}$& 24.91 &
					
					3.46 $_{\pm2.86}$&	11.4 $_{\pm3.01}$& 7.43
					\\ \hline
					DOC & \ding{55}  & 9.69 $_{\pm0.82}$&	4.75 $_{\pm0.82}$&	12.90 $_{\pm0.82}$&9.11	 &
					
					3.81 $_{\pm2.49}$&	14.37 $_{\pm1.28}$& 9.09
					\\ \hline
					DOC-C&\ding{55} &  11.37 $_{\pm3.81}$&	5.54 $_{\pm2.84}$&	11.09 $_{\pm3.91}$&  9.33 &
					
					3.50 $_{\pm2.64}$&	14.18 $_{\pm1.51}$& 8.84
					\\ \hline
					COT & \ding{55}  & 10.84 $_{\pm1.15}$&	7.31 $_{\pm3.86}$&	8.05 $_{\pm2.45}$&    8.73 &
					
					11.28 $_{\pm4.57}$&	 3.14 $_{\pm2.00}$& 7.21
					\\ \hline
					COT-C &\ding{55} & 12.36 $_{\pm3.72}$&	9.02 $_{\pm6.09}$&	7.42 $_{\pm3.12}$& 9.60 &
					
					9.43 $_{\pm4.03}$&	 3.25 $_{\pm1.95}$& 6.34
					\\ \hline
					Energy& \ding{55}  &  40.56 $_{\pm0.72}$&	47.46 $_{\pm18.83}$&	34.34 $_{\pm6.61}$& 40.79 & 
					
					4.44 $_{\pm4.20}$&	11.39 $_{\pm3.01}$& 7.92
					\\ \hline
					Energy-C &\ding{55}&  40.58 $_{\pm0.70}$&	47.82 $_{\pm18.53}$&	34.44 $_{\pm6.51}$& 40.95 & 
					
					4.34 $_{\pm4.17}$&	 11.39 $_{\pm3.01}$& 7.86
					\\\hline
					Nuclear Norm-C & \ding{55}  &2.29 $_{\pm4.25}$	&4.77 $_{\pm2.29}$&	17.86 $_{\pm4.92}$  & 8.31 & 		5.54  $_{\pm2.42}$	& 14.48  $_{\pm0.83}$ & 10.01  \\ \hline

     Agree Score & \ding{55} & 23.49	 & 	10.31	& 	22.17	& 18.66	 & 	
					6.2	&	8.06 & 7.13\\ \hline
					AC &  \checkmark & 	9.38& 		5.05& 		13.2& 		9.21 	& 
					13.46&	16.15 &14.81 \\ \hline
					
					Nuclear Norm & \checkmark  & 	0.05& 		4.48& 		20.4& 	8.31 & 	
					12.1	& 15.16 & 13.63 \\ \hline
					GradNorm & \checkmark  & 	27.3& 		7.82& 		7.45&  14.19 & 13.04 &	10.09 & 11.57	\\ 
					\midrule
					
					Our Approach &\checkmark & 8.77& 		3.64& 		1.39& 4.60	& 1.93&	3.29&	2.61 \\ \midrule
					GT Accuracy & & 41.59 & 81.46 &  50.67 & & 52.91 & 72.91 &\\
					\bottomrule
				\end{tabular}
			}
		}
	\end{center}
	\vspace{-6ex}
\end{table*}

The choice of the selected baseline methods is guided by specific criteria: They do not require any structural modifications to the network, avoid additional fine-tuning, and refrain from extensive data augmentation. These prerequisites align with the characteristics of our own method, ensuring a fair comparison.

We report two key metrics for each dataset: AE (absolute error between observed and predicted accuracy on the target data) and MAE (mean absolute error across different target sets). For baselines requiring source data during evaluation, we report performance with limited data availability, specifically using 1\% of source validation data (referred to as a 1\% inclusion ratio). 
Due to the sensitivity of sample-wise baselines to subset sample selection, we randomly sample 1\% of the data 20 times, reporting the mean and standard deviation for all methods across these samples. 
Unlike other baselines dependent on validation data, AgreeScore requires complete training data from the source domain to train a second classifier; we therefore use a 100\% inclusion ratio for that baseline.

Furthermore, we investigate the amount of data required for the existing source-based methods to match the performance of our method. To that end, we report the percentage of inclusion ratio needed for these baselines to reach our method's performance. Notably, many source-based methods underperform compared to our approach, even when using all available validation data.

\vspace{-2ex}
\subsection{Results}


We begin our exploration by examining single-domain datasets, followed by an in-depth analysis of multi-domain setups in the next section.

%
%

\noindent\textbf{Single-Source Setups}\label{single_source}

\begin{figure*}[!h]
	\centering
	\begin{subfigure}{0.32\textwidth}
		\includegraphics[width=\linewidth]{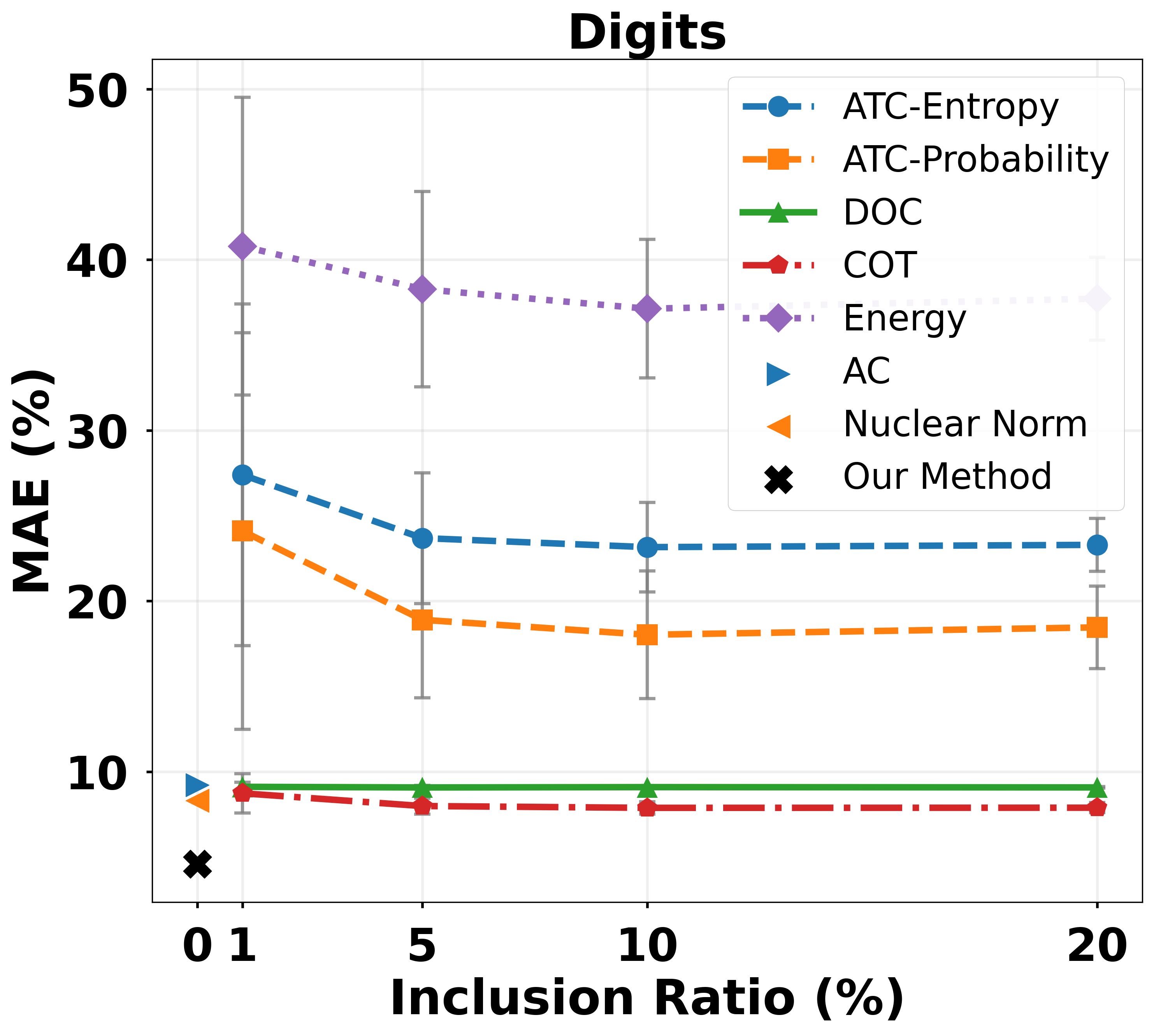}
	\end{subfigure}
	\begin{subfigure}{0.32\textwidth}
		\includegraphics[width=\linewidth]{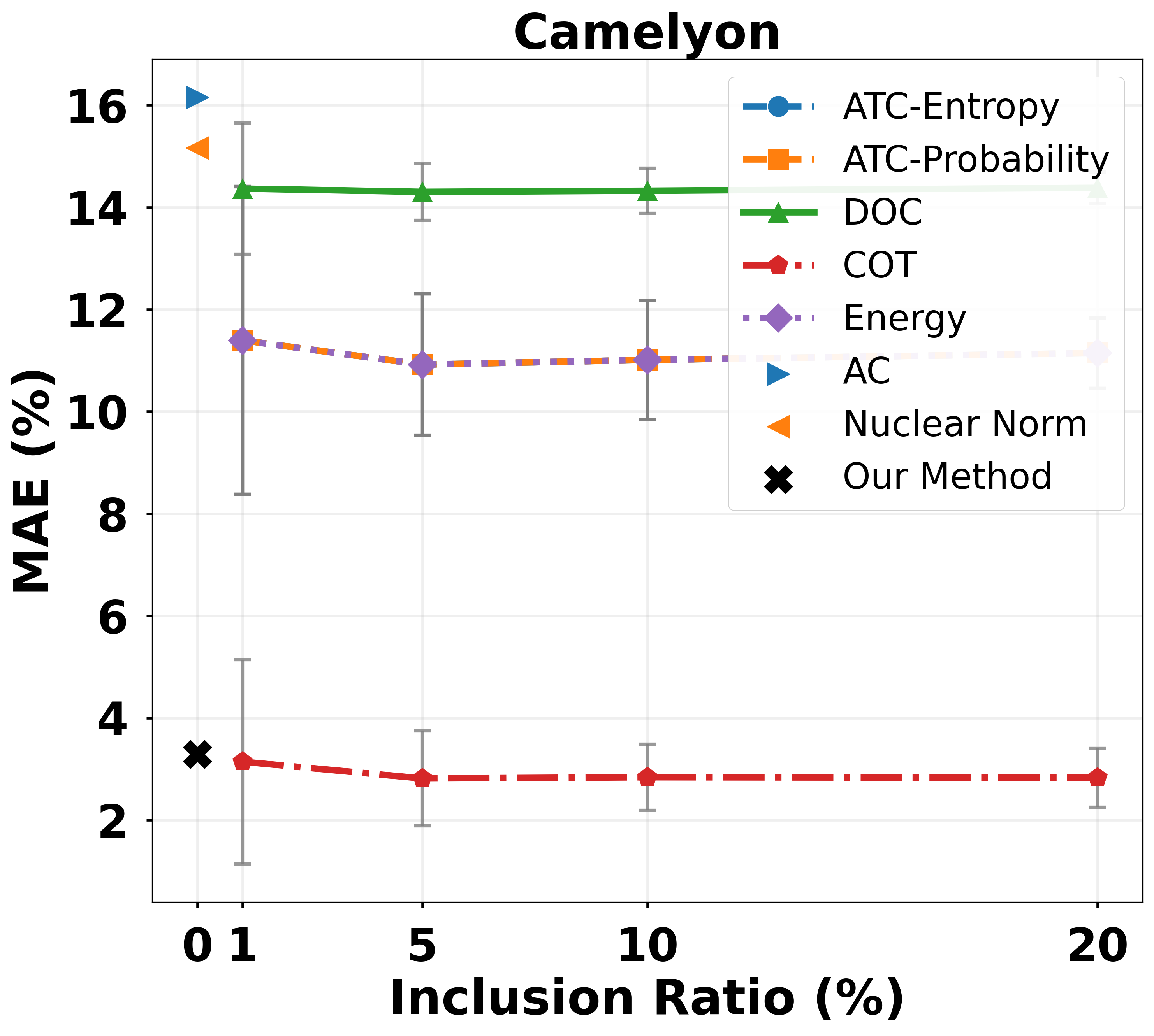}
	\end{subfigure}
	\begin{subfigure}{0.32\textwidth}
		\includegraphics[width=\linewidth]{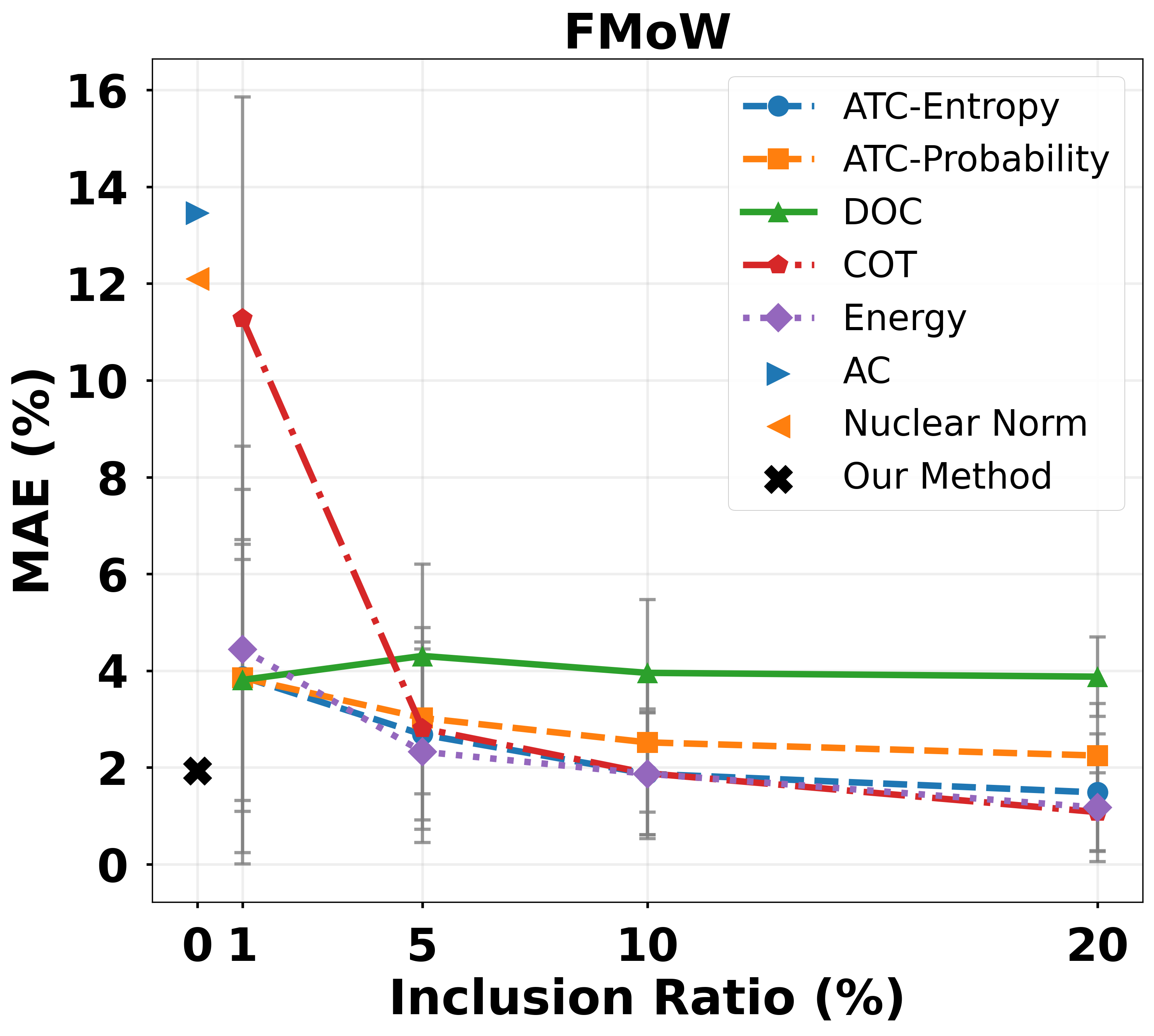}
	\end{subfigure}
	\vspace{-1ex}
	\caption{
		Mean Absolute Error Across Various Inclusion Ratios for Single-Domain datasets. Points represent the mean, vertical lines represent the standard deviation over 20 trials, each randomly selecting a percentage of source samples corresponding to the Inclusion Ratio.}
	\label{img:mae_vs_inclusion_single_domain}
\end{figure*}

Our experimental results are presented in Table \ref{table:single_source} and Figure \ref{img:mae_vs_inclusion_single_domain}. 

\begin{figure*}[!h]
	\centering
	\begin{subfigure}{0.95\textwidth}
		\includegraphics[width=\linewidth]{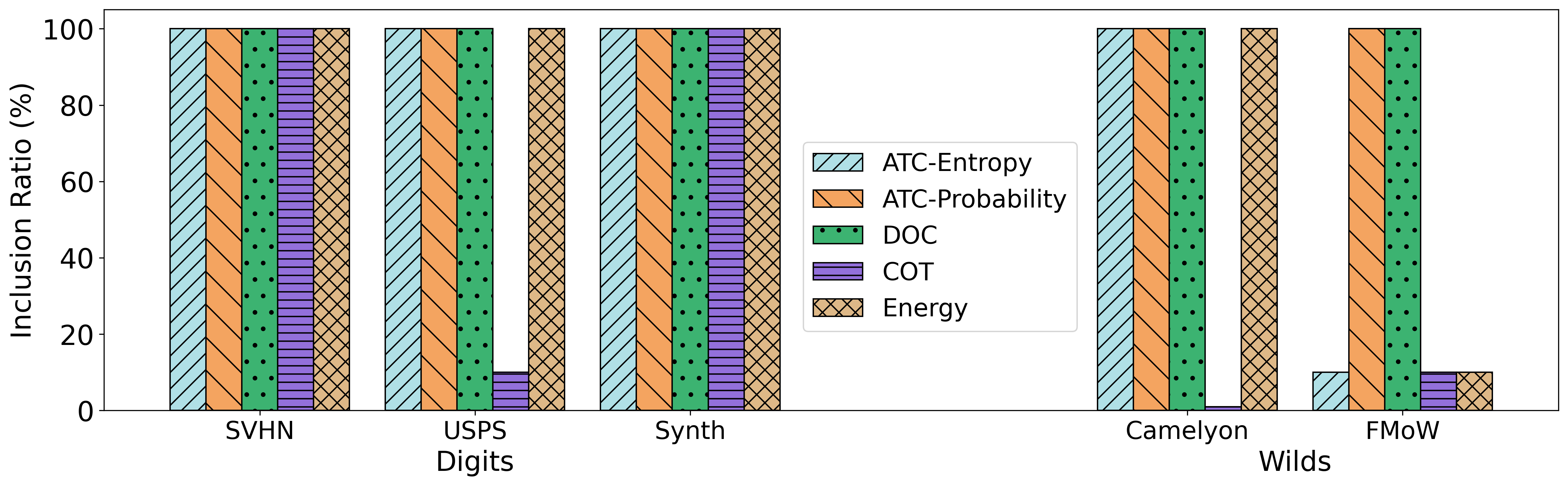}

	\end{subfigure}
	\vspace{-1ex}
	\caption{Inclusion Ratio needed for baselines to match our method's performance for Single-Domain datasets. A 100\% ratio indicates that baseline performance is inferior to ours, even when using all source data. }
	\label{img:inclusion_barplot_single_domain}

\vspace{-2ex}
\end{figure*}

We first observe that the performance of source-based baselines diminishes with a limited number of available source samples, which highlights the importance for source-free methods.
At the same time, the current source-free baselines exhibit poor performance. For example, in the Wilds benchmark, even with only 1\% of source data, all source-based methods surpass source-free approaches in performance.
 In contrast, our method outperforms both source-based and source-free baselines by a large margin, with just 2.61\% MAE on Wilds, and 4.6\% MAE on Digits. In comparison,  the nearest competitor in the Wilds setup is COT with 7.21\% MAE, and  in  Digits  setup - Nuclear Norm with  8.31\% MAE.

Our second observation is that calibration does not uniformly enhance source-based methods. In the  Digits setup, calibration degrades performance across all baseline methods; for the FMoW dataset, it enhences the DOC and COT methods but deteriorates ATC's performance. In contrast, for our method, the unsupervised calibration is paramount for achieving accurate performance predictions. 
By employing a probabilistic generative model, our approach adjusts the probabilities to align with the characteristics of the target dataset. Introducing this unsupervised calibration method allowed for a significant  performance  improvement compared to the standard GradNorm approach.

\begin{table*}[ht]
	\begin{center}
		\caption{
  Absolute Error between predicted and Ground Truth accuracy for \textbf{Multi-Source datasets}. Results for source-based baselines report the mean and standard deviation across 20 trials, with each trial randomly selecting 1\% of source samples.}
		\vspace{-2ex}
		\label{table:multi_domain}
		\scalebox{0.9}{
			{\renewcommand{\arraystretch}{1.2}%
				\begin{tabular}{@{}l|c|cccc|r@{}} 
					
					\toprule
					& Source-Free & PACS & VLCS & Office-Home & Terra Incognita  &\textbf{ MAE  }\\ 
					\midrule
					
					
					ATC$_{e}$ & \ding{55} &  12.19$_{\pm4.60}$ &	15.03$_{\pm5.15}$ & 10.48$_{\pm3.54}$ & 22.69$_{\pm6.60}$ & 15.10 \\ \hline
					ATC$_{e}$-C& \ding{55} &  15.84$_{\pm7.34}$	& 19.5$_{\pm5.95}$ & 10.82$_{\pm4.45}$	& 22.43$_{\pm5.67}$ & 17.15\\ \hline
					ATC$_{p}$ & \ding{55}& 12.28$_{\pm4.56}$	&14.87$_{\pm4.89}$	&11.02$_{\pm4.01}$	 &23.28$_{\pm5.85}$ & 15.36 \\ \hline
					ATC$_{p}$-C&  \ding{55}  &  12.47$_{\pm4.53}$	 &17.24$_{\pm5.68}$	&10.90$_{\pm4.11}$	&23.40$_{\pm6.27}$ & 16.00 \\ \hline
					DOC & \ding{55} & 11.03$_{\pm1.97}$	 &16.81$_{\pm4.90}$	&11.8$_{\pm3.31}$	&31.63$_{\pm3.02}$ & 17.82  \\ \hline
					DOC-C& \ding{55}  &  10.68$_{\pm2.34}$	 &17.4$_{\pm5.62}$	 &8.88$_{\pm3.62}$	&31.60$_{\pm3.69}$ & 17.14 \\ \hline
					COT & \ding{55}  & 22.87$_{\pm5.24}$&	24.12$_{\pm6.41}$&	48.34$_{\pm3.32}$&	11.54$_{\pm4.40}$& 26.72 \\ \hline
					COT-C& \ding{55} &  22.58$_{\pm4.84}$&	19.73$_{\pm5.76}$&	42.22$_{\pm2.17}$&	 11.25$_{\pm4.43}$& 23.95 \\ \hline
					Energy& \ding{55}&  12.40$_{\pm4.62}$&	20.58$_{\pm5.23}$&	11.56$_{\pm4.70}$&	20.83$_{\pm6.92}$& 16.34 \\ \hline
					Energy-C& \ding{55}  &  12.53$_{\pm4.72}$&	 22.29$_{\pm5.33}$&	 11.38$_{\pm4.87}$&	20.74$_{\pm7.09}$& 16.74 \\ \hline
					Nuclear Norm-C &\ding{55} & 9.66 $_{\pm1.7}$&		9.75 $_{\pm2.92}$&		10.93 $_{\pm2.6}$ &	17.92 $_{\pm2.18}$ & 12.07\\ \hline
					Agree Score &\ding{55} & 3.65	&	10.27 &	2.34&	15.61& 7.97 \\ \hline
					AC &\checkmark & 					12.82&	20.15&	22.1&	33.88& 22.24 \\ \hline

					Nuclear Norm &\checkmark & 			11.11&	11.84&	21.72&	19.12& 15.95 \\ \hline
					GradNorm &\checkmark& 				14.42&	21.30&	24.11&	36.55& 24.10 \\ 
					\midrule
					
					Our Approach &\checkmark & 4.27&	5.75&	7.08&	8.72& 6.46  \\ \hline
					GT Accuracy & & 82.00 & 74.83 & 63.31 & 50.07 & \\
					\bottomrule
				\end{tabular}
				
			}
		}
	\end{center}
	\vspace{-3ex}
\end{table*}

Next, we explore how many samples are required for the baseline methods to match our performance.  Remarkably, for the SVHN and SYNTH datasets, our method surpasses all source-based methods, even with full source data availability, as illustrated in Figure 3. For other datasets, the COT baseline improves as more data becomes available, equalling our method's performance with a 10\% inclusion ratio for USPS and FMoW, and 1\% for Camelyon. Nonetheless, at lower inclusion ratios, COT and other source-based methods exhibit high variance in MAE, indicating instability. Figure 4 shows that while COT achieves less than 2\% MAE at a 10\% inclusion ratio, reducing this ratio to 1\% causes a significant spike in MAE to 11.23\%, with a standard deviation of 4.57\%.
In contrast, our approach is fully deterministic and thus maintains consistent and stable performance, even in the absence of source data.


\noindent\textbf{Multi-Domain Setups}

In the following section,  we transition to multi-domain setups. The outcomes of our experiments are detailed in Table \ref{table:multi_domain} and Figure \ref{img:inclusion_barplot_multi_domain}.

Contrary to the previous setting in which a single domain was designated as the test set, the current datasets comprise four distinct domains. In each configuration, each domain is alternately used as the test set, with the remaining three domains employed for training. As a result, in this scenario, we evaluate the performance of four different models for each dataset, as opposed to the earlier setup where only one model was associated with each dataset.

 \begin{figure*}[!h]
	\centering
	\begin{subfigure}{0.24\textwidth}
		\includegraphics[width=\linewidth]{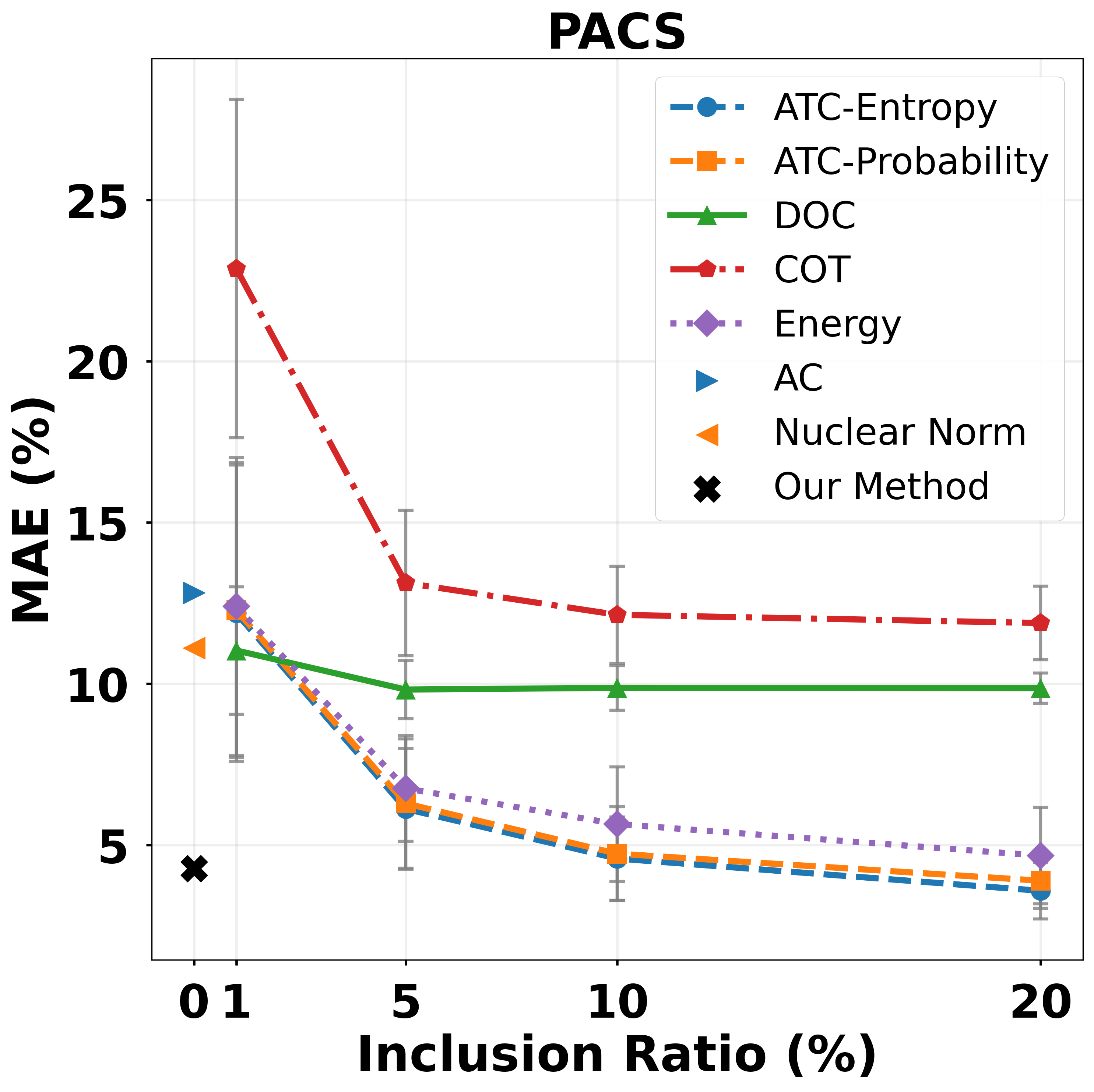}
	\end{subfigure}
	\begin{subfigure}{0.24\textwidth}
		\includegraphics[width=\linewidth]{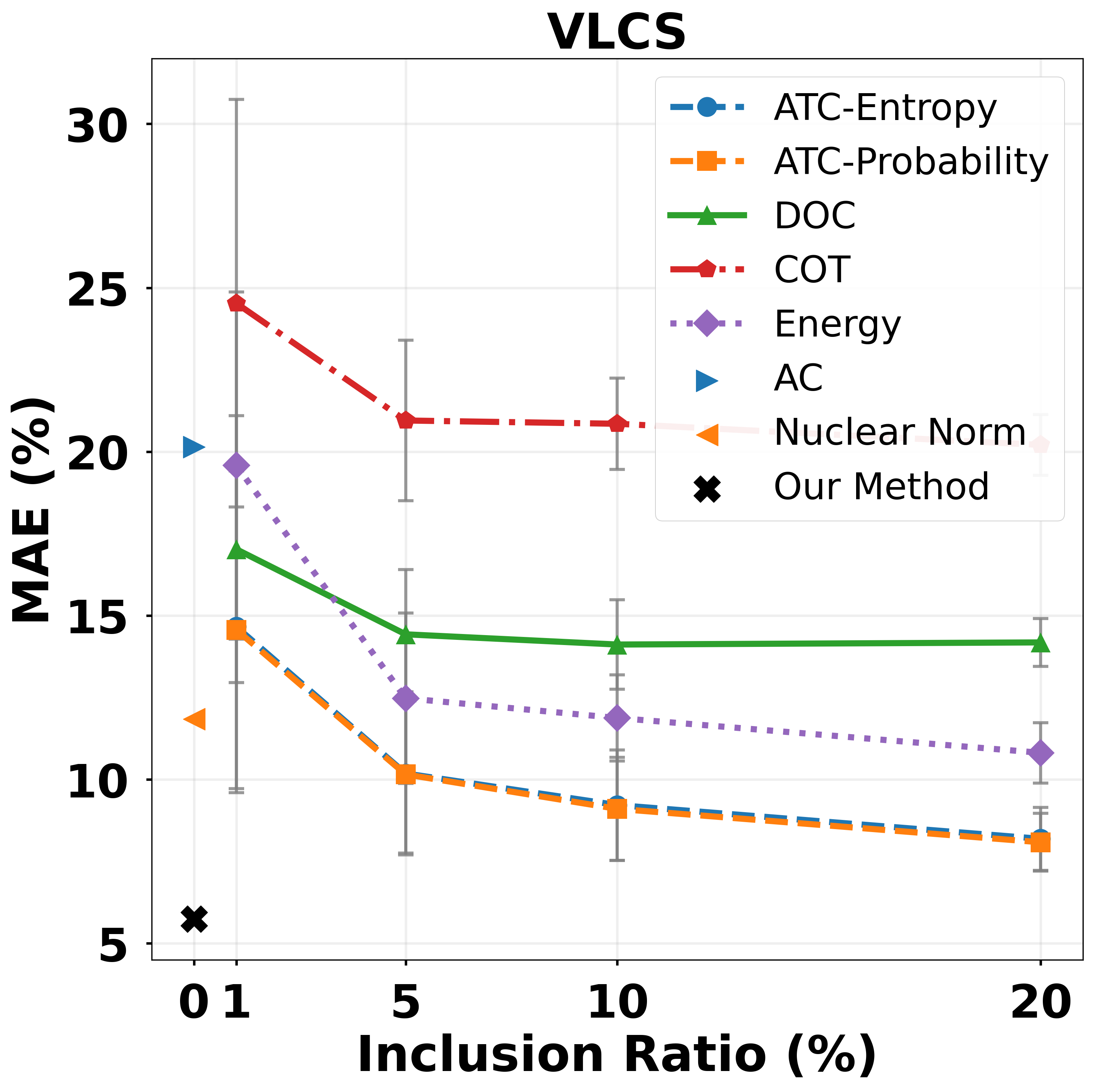}
	\end{subfigure}
	\begin{subfigure}{0.24\textwidth}
		\includegraphics[width=\linewidth]{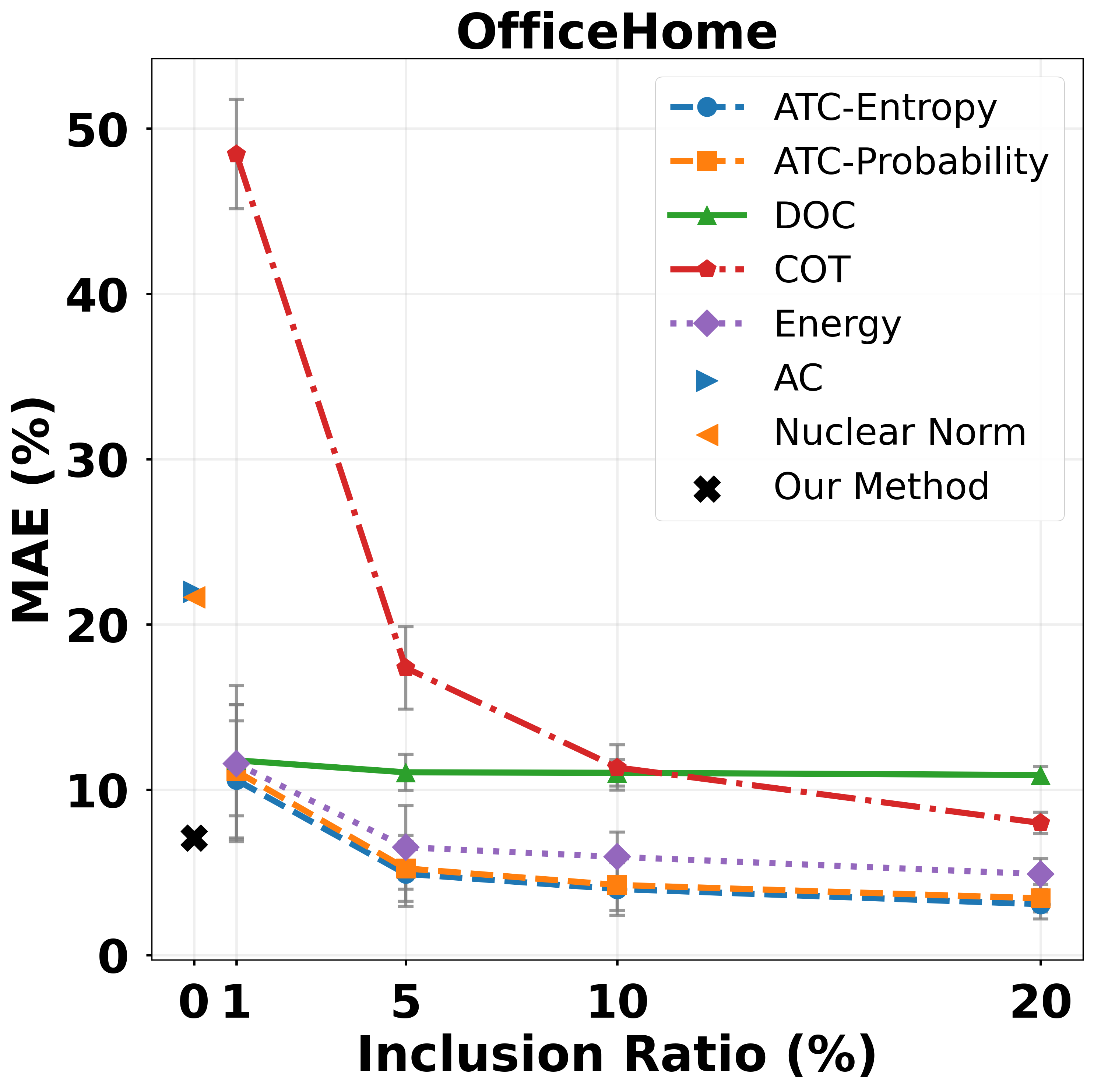}
	\end{subfigure}
	\begin{subfigure}{0.24\textwidth}
		\includegraphics[width=\linewidth]{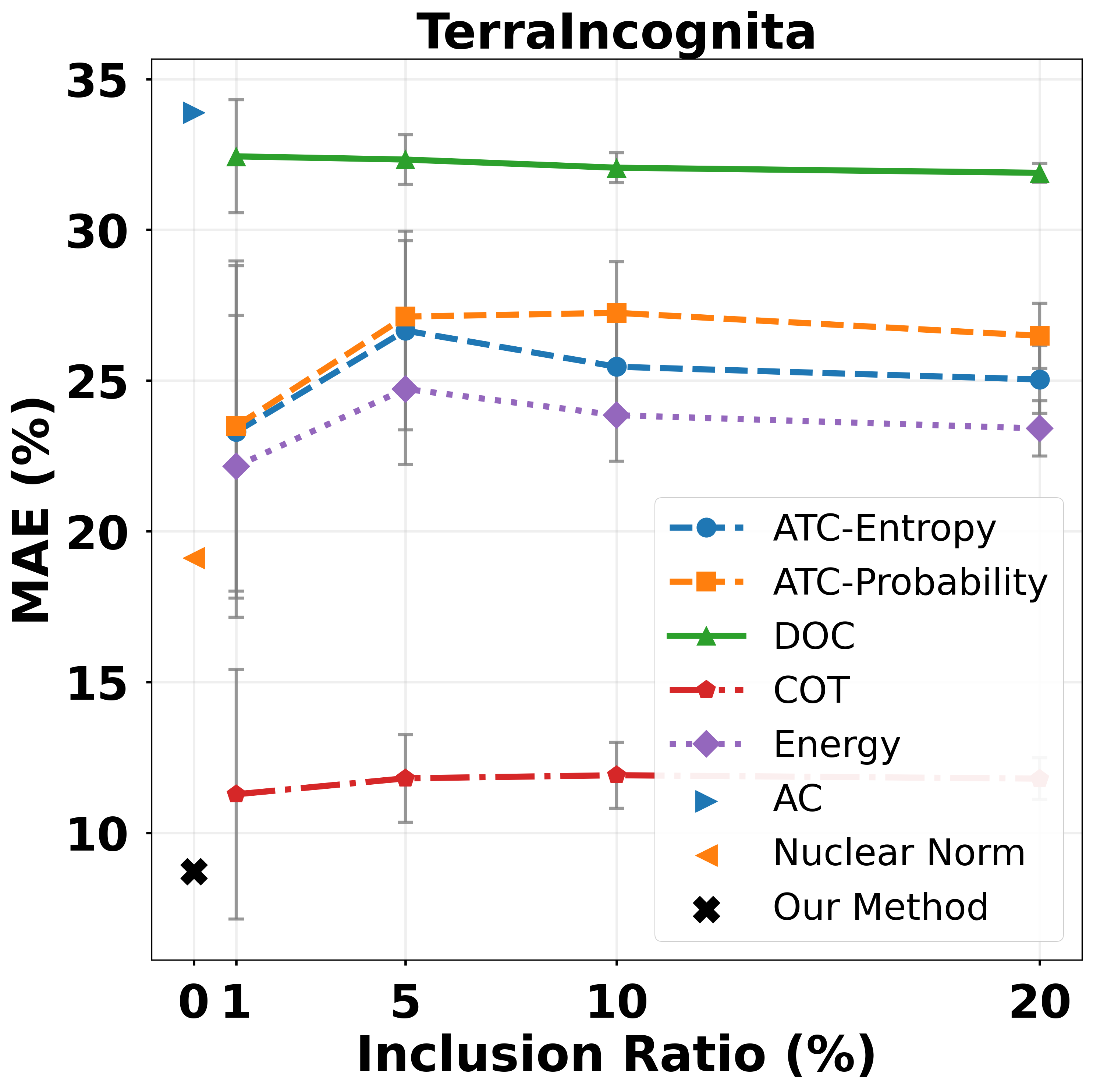}
	\end{subfigure}
	\vspace{-1ex}
	\caption{	Mean Absolute Error Across Various Inclusion Ratios for Multi-Domain datasets. Points represent the mean, and vertical lines represent the standard deviation over 20 trials, each randomly selecting a percentage of source samples corresponding to the Inclusion Ratio.}
	\label{img:mae_multi_domain}
\vspace{-1ex}
\end{figure*}

\begin{figure*}[!h]
	\centering
	\begin{subfigure}{0.99\textwidth}
		\includegraphics[width=\linewidth]{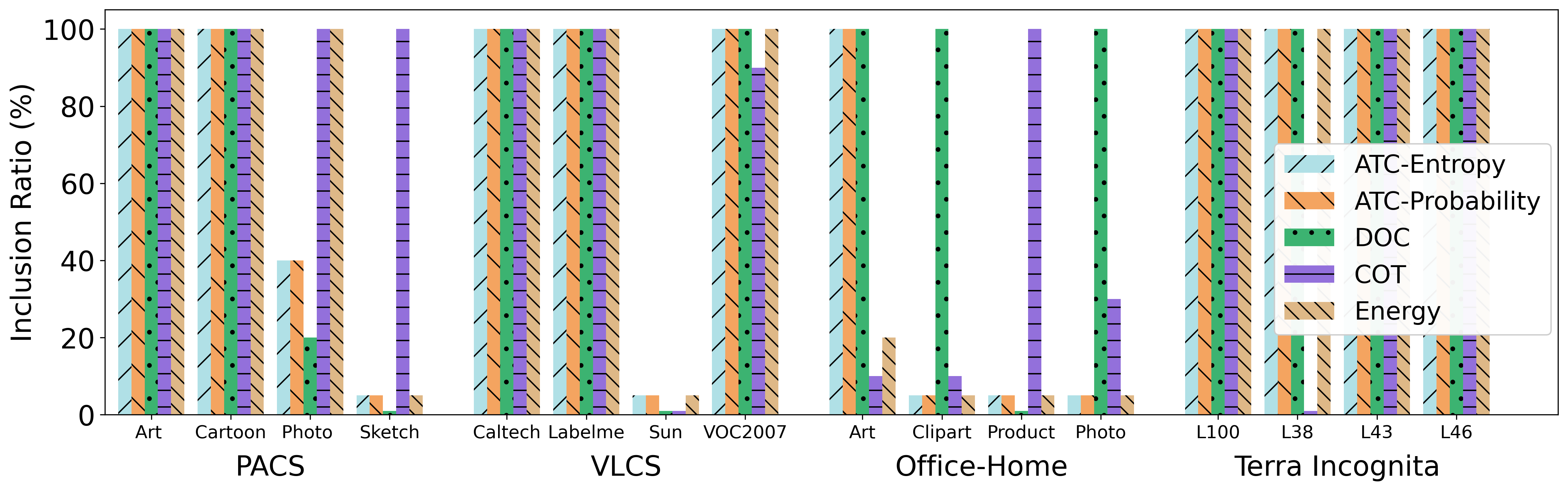}

	\end{subfigure}
	\vspace{-2ex}
	\caption{Inclusion Ratio needed for baselines to match our method's performance for Multi-Domain datasets. A 100\% ratio indicates that baseline performance is inferior to ours, even when using all source data.}
	\label{img:inclusion_barplot_multi_domain}
	\vspace{-3ex}
\end{figure*}

Our initial observation reveals that the COT baseline, which exhibited superior performance in the single-domain setup, significantly underperforms in the multi-domain setup. Its performance is, on average, even inferior to that of source-free baselines, with an average MAE of 26.72\%.
  Baselines used on PACS and VLCS benifit from more data to achieve competitve results (see Figure \ref{img:mae_multi_domain}). This contrasts with the Terra Incognita dataset, where an increase in source data does not correspond to enhanced performance across baselines. Specifically, Figure \ref{img:mae_multi_domain} demonstrates that the MAE does not diminish with an increased inclusion ratio; in fact, it escalates for several baselines, including ATC, COT, and Energy. This outcome highlights the heterogeneity of the source dataset, which includes datasets from diverse domains.

 In contrast, our proposed method consistently outperforms all other baselines, especially when little to no source data is available, as evidenced in Table~\ref{table:multi_domain}. Specifically, on the Office Home dataset, our method surpasses all source-free baselines by a margin of at least 14\% MAE. Additionally, most baseline methods on that dataset only begin to macth the performance of our approach at 10\% inclusion ratio. Remarkably, across all test sets of Terra Incognita, our method maintains superiority over the baselines even at a 100\% inclusion ratio, as depicted in Figure \ref{img:inclusion_barplot_multi_domain}. This trend also extends to the Art and Cartoon test sets of PACS and the Caltech and LabelMe datasets of VLCS, demonstrating the robustness and adaptability of our method across diverse domains.




\subsection{Ablation Study and Limitations}

\begin{wrapfigure}{l}{0.44\textwidth} 
	\vspace{-5ex}
	\centering
	\includegraphics[width=1.\linewidth]{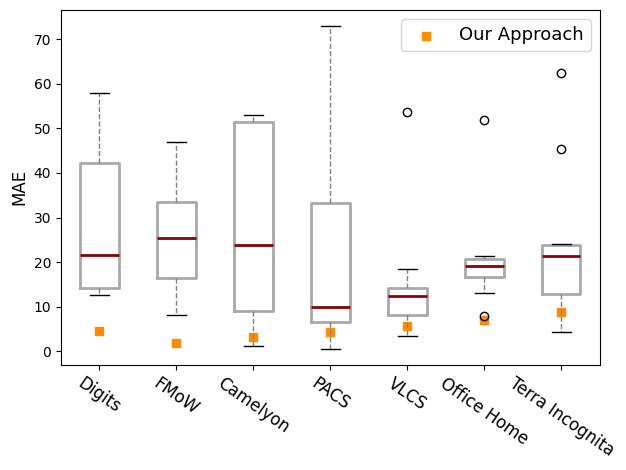}
 \caption{MAE Variance (sensitivity) of GradNorm to temperature scaling averaged on all datasets. The MAE for our approach is shown in Orange.}
	\label{img_tempr}
 \vspace{-5ex}
\end{wrapfigure}


\textbf{Ablation study}. In this section, we investigate the viability of our approach without the proposed calibration. This involves the use of GradNorm with the softmax function, parameterized by a constant temperature. This experiment aims to assess the effectiveness of scaling the softmax function with a fixed temperature, specifically to examine whether there exist an optimal threshold across all target datasets.

The results indicate that including the temperature parameter consistently improves the predictor's performance. This suggests that the original predictions for all datasets were overly confident. However, the optimal temperature value varies among datasets. 
\katyatodo{For instance, for the Camelyon17 dataset, the best threshold value is 0.7, resulting in a 3.6\% MAE, while the same threshold for the Digits datasets yields a 14.2\% MAE.}

In Figure~\ref{img_tempr}, we further demonstrate the MAE variance of the GradNorm-based performance estimator across different temperatures. The large variance in MAE values highlights that accurate selection of temperature is critical for the GradNorm estimator. Without access to labeled test data, this choice cannot be made accurately. Therefore, our introduced unsupervised calibration method is crucial for obtaining reliable estimations.

\katyatodo{
For additional experiments, including comparisons of computational efficiency between methods and the effectiveness of our method versus dataset-wide methods, please refer to the Supplementary Material.
}

\noindent\textbf{Limitations.} \katyatodo{Our method relies on the assumption that the logits follow a Gaussian distribution, as used in several other studies~\cite{matthews2018gaussian, garriga-alonso2018deep}. Also, while our method can be sensitive to sparse datasets, making the accurate estimation  of Gaussian parameters challenging with few samples per
 class, we address this by adopting a zero mean and unit variance approach, which works well in practice.}

\vspace{-1ex}
\section{Conclusion}
\vspace{-0.5ex}

In conclusion, our work addresses the challenge of accurately estimating model performance in a source-free and domain-invariant setting, making it relevant for real-world scenarios where the source data is unavailable. Our novel unsupervised approach utilizes a generative model to reduce prediction certainty, addressing the issue of neural networks being overly confident. 
Then, the prediction correctness is evaluated using gradient norm derived from associated loss. Extensive experiments on diverse benchmark datasets demonstrate our approach's significant performance improvement over current state-of-the-art methods in source-free domain-invariant accuracy prediction.

\section*{Acknowledgements}
This research is supported by the National Key Research and Development Program of China No. 2020AAA0109400 and the Shenyang Science and Technology Plan Fund (No. 21-102-0-09). This work is also partially supported by Australian Research Council Project FT230100426.

	%
	%
	\bibliographystyle{splncs04}
	\bibliography{main}

	\title{Source-Free Domain-Invariant \\ Performance Prediction \\ 
	{\large	 Supplementary Material}} 
	
	\titlerunning{Source-Free Domain-Invariant Performance Prediction}
	
	\author{Ekaterina Khramtsova\inst{1}\orcidlink{0000-0001-7531-4491} \and
		Mahsa Baktashmotlagh\inst{1}\orcidlink{0000-0001-5255-8194} \and
		Guido Zuccon\inst{1}\orcidlink{0000-0003-0271-5563}\and\\
		Xi Wang\inst{2}\orcidlink{0009-0002-1724-7694}\and
		Mathieu Salzmann\inst{3}\orcidlink{0000-0002-8347-8637}
  }
	
	\authorrunning{E.Khramtsova et al.}
	
	\institute{
        The University of Queensland, Australia \and 
        Neusoft, China \and 
        École Polytechnique Fédérale de Lausanne (EPFL), Switzerland \\
        \email{e.khramtsova@uq.edu.au}\\        \url{https://github.com/khramtsova/source_free_pp/}
        }
	\maketitle	
	\begin{figure}[ht]
    \begin{center}
    \vspace{-3ex}
      \begin{subfigure}{0.3\textwidth}
            \includegraphics[width=\linewidth]{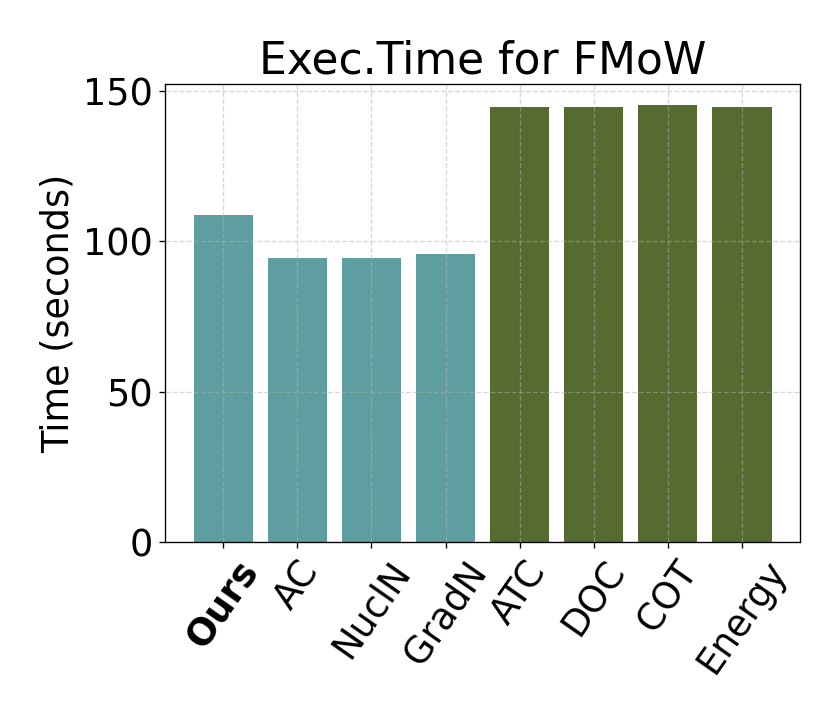}
     \end{subfigure}
     \begin{subfigure}{0.3\textwidth}
         \includegraphics[width=\linewidth]{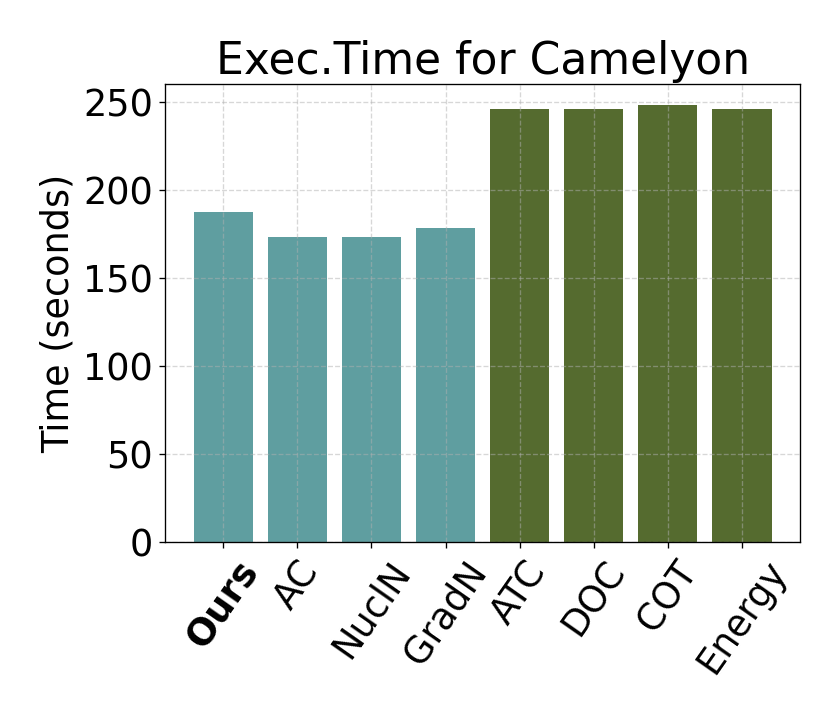}
  \end{subfigure}
  \begin{subfigure}{0.3\textwidth}
            \includegraphics[width=\linewidth]{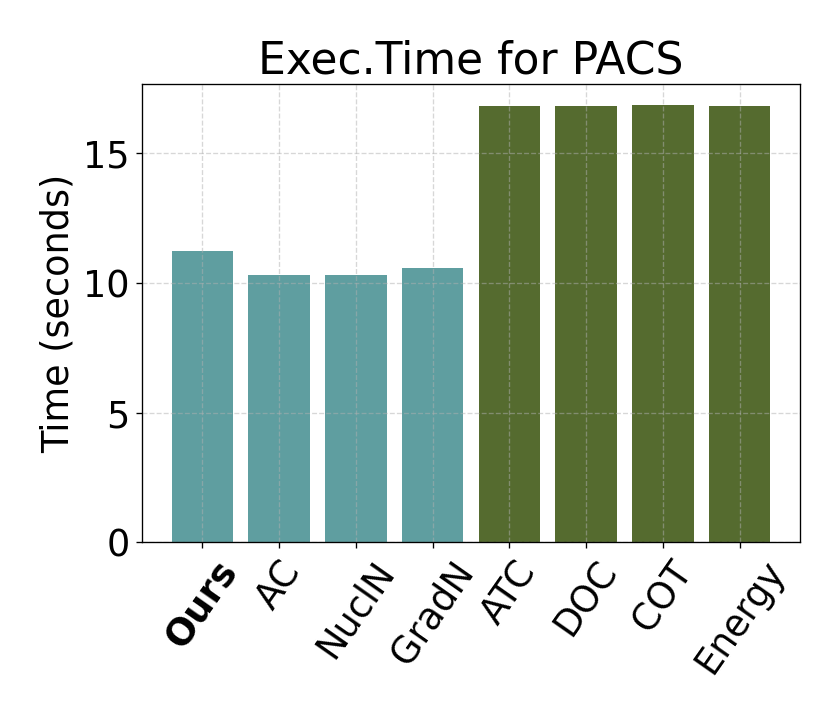}
     \end{subfigure}
  
  \vspace{-1ex}
     \begin{subfigure}{0.3\textwidth}
            \includegraphics[width=\linewidth]{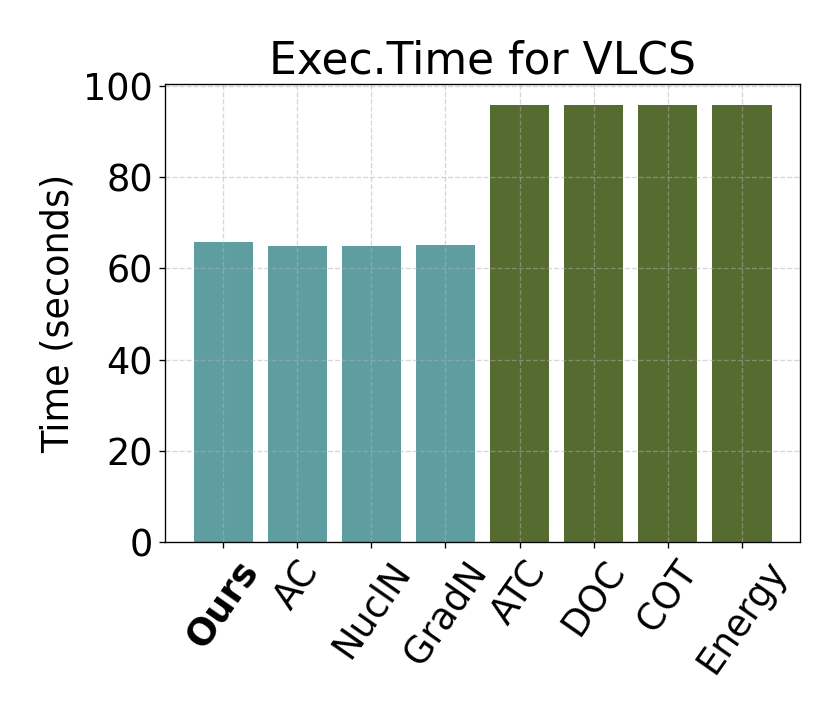}
     \end{subfigure}
     \begin{subfigure}{0.3\textwidth}
         \includegraphics[width=\linewidth]{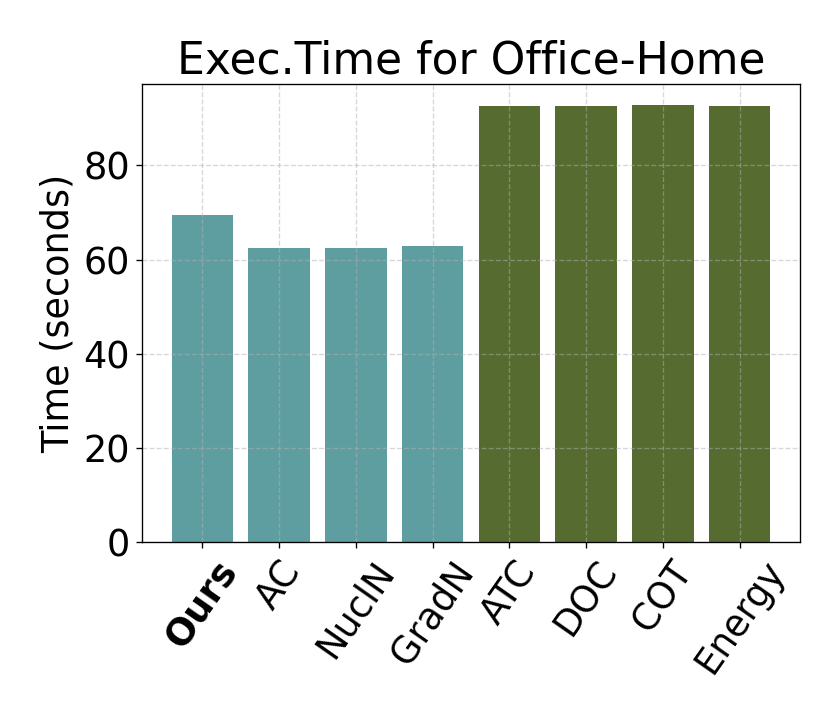}
  \end{subfigure}  
     \begin{subfigure}{0.3\textwidth}
         \includegraphics[width=\linewidth]{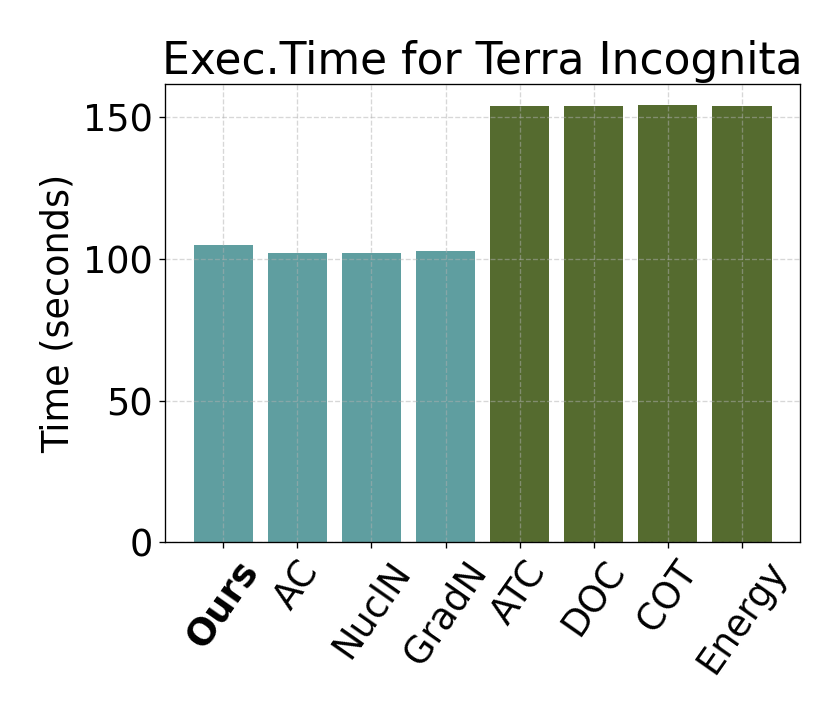}
  \end{subfigure}
 \vspace{-2ex}
         \caption{Execution Time. Source-Free (Blue), Source-Based (Green)}
         \label{fig:time_comparison}
 \vspace{-4ex}
 \end{center}
 \end{figure}


\katyatodo{This supplementary material consists of two main parts. The first part discusses the efficiency of various sample-based methods used in the main paper. The second part explores dataset-wide methods, which were not covered in the main paper due to their fundamentally different experimental assumptions.}

\vspace{-1ex}
\section*{Efficiency Analysis of Sample-Based Methods}
\vspace{-1ex}

 \begin{table}[!h]
    \caption{Complexity Comparison wrt Network Pass Requirement}
    \vspace{-4ex}
    \label{table:forw_backw}
    \begin{center}
    \resizebox{0.7\linewidth}{!}{
    {\renewcommand{\arraystretch}{1.2}%
    \begin{tabular}{c|c|c|c|c}
    
    \toprule
    
     \multicolumn{1}{c|}{ }&\multicolumn{2}{|c|}{ Source} &\multicolumn{2}{|c}{Target}\\
     
      \hline
        & Forward & Backward & Forward & Backward \\
     \midrule   
   AC, Nuclear Norm &   \checkmark & \ding{55} &\ding{55} & \ding{55}\\ \hline
    ATC, DOC, COT, Energy & \checkmark & \ding{55} & \checkmark & \ding{55} \\ \hline
    Agree Score & \checkmark \checkmark & \checkmark\checkmark & \checkmark & \checkmark \\ \hline
    Our Approach, GradNorm & \ding{55} & \ding{55} & \checkmark &\checkmark \\ 
    
    \bottomrule
    \end{tabular}
    }
    }
    \end{center}
    \vspace{-1ex}
\end{table}

	\begin{figure*}[!h]
		\vspace{-3ex}
			\begin{subfigure}{0.4\textwidth}
			\includegraphics[width=\linewidth]{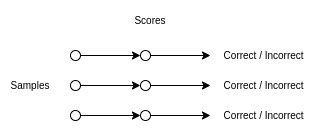}
			\vspace{3ex}
			\caption{Sample-wise methods}
			\label{img:sample_wise}
		\end{subfigure}
		\hfill
		\begin{subfigure}{0.55\textwidth}
			\includegraphics[width=\linewidth]{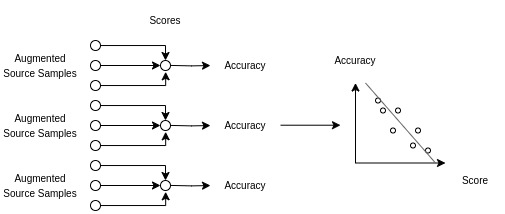}
			\caption{Dataset-wide methods}
			\label{img:dataset_wide}
		\end{subfigure}
		\caption{Conceptual difference between Sample-wise and Dataset-wide methods}
		\label{img:exp_setups}
		\vspace{-3ex}
	\end{figure*}

\katyatodo{
The computational complexity of each method is determined by the number of forward and backward passes of the dataset through the network. While our method requires a backward pass through the network's last layer, all source-based approaches involve a forward pass of the source data. With a 100\% inclusion rate, this takes more time than a backward pass, as illustrated in Figure~\ref{fig:time_comparison}. Additionally, the Agree Score requires an additional full training on the source data. Consequently, only source-free methods are more computationally efficient than our approach; however, they are significantly less effective. The backward and forward passes required for each method are summarized in Table~\ref{table:forw_backw}.}

\katyatodo{
 Note that pseudo-labeling does not introduce any  computational overhead:  it is performed by forwarding the data through the network under evaluation and considering the most probable prediction of that network as the label.}
 

\vspace{-1ex}
\section*{Exploration of Dataset-wide Methods}
\vspace{-1ex}
 
 In this section, we extend our analysis to  include  additional performance prediction methods, providing an alternative viewpoint on the performance prediction problem. We outline the primary differences and challenges associated with this alternative perspective in Section~\ref{sec:method_compar}, followed by the presentation of experimental results in Section~\ref{sec:experimental_res} and its analysis in Section~\ref{sec:performace_analysis}.

 \vspace{-2ex}
 \subsection{Sample-wise vs. Dataset-wide methods}\label{sec:method_compar}
	This study introduces a novel approach for unsupervised source-free performance prediction. 
	All the considered baselines represent Sample-wise methods (see Fig.\ref{img:sample_wise}) and share a common structure. In essence, each sample is forwarded through the network and  assigned a score based on either the network's output, or the network's parameters (e.g., the gradient, as per our method).  These scores are further used to predict the overall accuracy, for example by taking the average confidence across all samples (AC\cite{AC}). 
	Other methods, including ours, extend this by predicting the correctness of each sample based on specified criteria.  For instance, in the ATC method~\cite{ATC}, a sample is predicted as correct if its score is below a certain threshold, estimated from the source set. Our method considers a sample to be predicted correctly if the gradient of its most probable prediction is smaller than that towards a uniform prediction.

	\begin{table*}[ht]
		\begin{center}
			\caption{
				Absolute Error between predicted and Ground Truth accuracy. Results for \textbf{Dataset-wide baselines} report the mean and standard deviation across 3 trials, with each trial randomly selecting \textbf{1\%} of source samples. 
			}
			\vspace{-1ex}
			\label{table:dataset_wide_1per}
			\scalebox{0.8}{
				{\renewcommand{\arraystretch}{1.2}%
					\begin{tabular}{@{}l|c|clcccc|r@{}} 
						
						\toprule
						& Type & FMoW & Camelyon & PACS & VLCS & Office-Home & Terra Incognita  &\textbf{ MAE  }\\ 
						\midrule
						GT Accuracy &  & 52.91 &72.91 & 82.00 & 74.83 & 63.31 & 50.07 & \\ \midrule
						
						AC & S-W & 13.46 & 16.15 & 	12.82&	20.15&	22.1&	33.88& 19.76  \\
						
						Nuclear Norm &S-W & 12.1 & 15.16& 	11.11&	11.84&	21.72&	19.12& 15.17 \\
						
						GradNorm & S-W & 13.04& 10.09 &  14.42&	21.30&	24.11&	36.55& 19.92 \\  \hline
						
						AC & D-W & 3.8  $_{\pm2.09}$ &  8.8 $_{\pm2.67}$ & 11.11 $_{\pm2.19}$&	18.8 $_{\pm6.64}$	&10.59 $_{\pm3.56}$&	23.57 $_{\pm2.09}$ &  13.57\\ 
						Nuclear Norm &D-W & 4.2 $_{\pm2.7}$ &5.78 $_{\pm1.8}$ & 10.43 $_{\pm2.65}$&	21.28 $_{\pm13.74}$ &	11.07 $_{\pm4.58}$& 	8.68 $_{\pm3.35}$ & 10.24 \\ 
						GradNorm & D-W & 2.0 $_{\pm0.82}$ 	&  8.51  $_{\pm1.37}$   &11.33 $_{\pm2.72}$&	19.51 $_{\pm6.61}$	&11.66 $_{\pm3.76}$ & 19.95 $_{\pm2.56}$ & 12.16 \\ 
						
						OT cost & D-W &  8.99 $_{\pm2.37}$	&2.91 $_{\pm0.27}$ &				25.9 $_{\pm5.02}$&	22.69 $_{\pm3.29}$&	44.57 $_{\pm5.02}$&	13.31 $_{\pm2.39}$& 19.73 \\
						Energy &  D-W &  5.7 $_{\pm2.86}$  & 11.57 $_{\pm2.19}$ &					 11.0 $_{\pm3.88}$ & 	17.81 $_{\pm8.25}$ &	9.73 $_{\pm2.5}$ &	23.84 $_{\pm2.02}$ & 13.27\\
						FID & D-W & 13.53 $_{\pm3.75}$ 		 &0.47 $_{\pm0.4}$ &			31.17 $_{\pm11.25}$ &	28.52 $_{\pm16.78}$ &	40.12 $_{\pm18.32}$ &15.80 $_{\pm3.06}$  & 21.60 \\
						Dispersion & D-W & 4.84 $_{\pm3.2}$		& 4.67 $_{\pm1.76}$ &			 13.17 $_{\pm4.8}$&	20.1 $_{\pm7.55}$& 	11.49 $_{\pm3.29}$ &	31.30 $_{\pm2.64}$ & 14.26 \\
						\midrule
						
						Our Approach &S-W & 1.93 & 3.29 & 4.27&	5.75&	7.08&	8.72& 5.17  \\
						\bottomrule
					\end{tabular}
					
				}
			}
		\end{center}
		\vspace{-5ex}
	\end{table*}

	However, we distinguish another class of performance prediction methods, namely Dataset-wide methods (see Fig.\ref{img:dataset_wide}) , which we excluded from the main body of the paper due to their significantly different computational complexity and experimental setup.  

	The main difference of Dataset-wide methods is that they assign a single score to the entire dataset instead of individual samples, requiring learning the correlation between this score and dataset accuracy, typically through Linear Regression. This approach requires generating training data with varied accuracies, often by using corrupted or augmented versions of the source dataset. The trained model then predicts performance based on the dataset's score.

	In contrast, our method predicts performance with a single pass of the test data through the network, whereas Dataset-wide methods need augmented data versions and multiple network passes for training. With most existing methods requiring at least 500 training data points, this translates to 500 additional network passes with source validation data. Despite the computational overhead, we adapted Dataset-wide methods to our setup for experimental completeness.

\subsection{Experimental Results}\label{sec:experimental_res}

We adapted the following existing Sample-wide baselines: AC~\cite{AC}, OT Cost~\cite{lu2023cot_for_ood}, NuclearNorm~\cite{Deng2023ConfidenceAD}, and Energy~\cite{peng2024energybased}. To convert them into Dataset-wide methods, we calculated the average corresponding score across the dataset. In addition,
we include the following Dataset-wide baselines:	GradNorm~\cite{xie2024leveraging}, FID~\cite{deng2020labels}, Dispersion Score~\cite{xie2023on}. Note that differently from our setup, where we rely on the gradient of each sample, the Dataset-wide variant by Xie et al.~\cite{xie2024leveraging} uses dataset-level gradient.

	The results of the experiment are presented in  Table \ref{table:dataset_wide_1per} and Table \ref{table:dataset_wide_5per} for 1\% and 5\% openness, respectively.

\begin{table*}[ht]
	\begin{center}
		\caption{
			Absolute Error between predicted and Ground Truth accuracy. Results for \textbf{Dataset-wide baselines} report the mean and standard deviation across 3 trials, with each trial randomly selecting \textbf{5\%} of source samples. 
		}
		\label{table:dataset_wide_5per}
		\scalebox{0.8}{
			{\renewcommand{\arraystretch}{1.2}%
				\begin{tabular}{@{}l|c|clcccc|r@{}} 
					
					\toprule
					& Type & FMoW & Camelyon & PACS & VLCS & Office-Home & Terra Incognita  &\textbf{ MAE  }\\ 
					\midrule
					GT Accuracy & & 52.91 & 72.91 & 82.00 & 74.83 & 63.31 & 50.07 & \\ \midrule
					
					AC & S-W & 13.46 & 16.15 & 	12.82&	20.15&	22.1&	33.88& 19.76  \\
					
					NuclearNorm &S-W & 12.1 & 15.16& 	11.11&	11.84&	21.72&	19.12& 15.17 \\
					
					GradNorm & S-W & 13.04& 10.09 &  14.42&	21.30&	24.11&	36.55& 19.92 \\  \hline
					
					AC & D-W & 4.12 $_{\pm0.88}$ & 8.48  $_{\pm0.44}$ &					8.65 $_{\pm1.23}$& 	12.92 $_{\pm3.28}$& 	7.55 $_{\pm1.49}$ & 22.98 $_{\pm1.33}$ & 10.78\\ 
					NuclearNorm &D-W & 3.67 $_{\pm1.89}$		&4.16  $_{\pm0.47}$ 	 &		 6.32 $_{\pm2.38}$&	11.84 $_{\pm3.76}$&	5.57 $_{\pm3.01}$ &8.16 $_{\pm1.27}$  & 6.62 \\ 
					GradNorm & D-W & 3.1  $_{\pm1.86}$	& 5.81 $_{\pm0.17}$ &	10.61  $_{\pm1.31}$		&14.28  $_{\pm3.18}$ &9.61  $_{\pm1.04}$ & 	17.32  $_{\pm1.92}$ & 10.12\\ 
					
					OT cost & D-W & 0.91 $_{\pm0.73}$				& 	3.3  $_{\pm0.49}$ & 15.34 $_{\pm2.47}$&	14.01 $_{\pm3.57}$ 	& 16.41 $_{\pm2.0}$  & 12.81 $_{\pm0.9}$& 10.46 \\
					Energy &  D-W & 5.65 $_{\pm1.59}$				&	10.21  $_{\pm0.15}$  &9.21 $_{\pm2.11}$&	14.63 $_{\pm3.96}$&	9.51 $_{\pm2.95}$ & 22.48 $_{\pm1.17}$& 11.95 \\
					FID & D-W & 2.41 $_{\pm1.9}$				& 	0.68  $_{\pm0.57}$ & 14.0 $_{\pm6.05}$	& 18.31 $_{\pm7.57}$	& 19.12 $_{\pm3.39}$ &  11.36 $_{\pm3.38}$ & 10.98 \\
					Dispersion & D-W & 5.41 $_{\pm2.66}$	 		& 3.07  $_{\pm0.64}$ 	&	8.45 $_{\pm2.57}$&	13.32 $_{\pm3.52}$&	14.58 $_{\pm3.63}$ &25.0 $_{\pm0.88}$ & 11.64\\
					\midrule
					
					Our Approach &S-W & 1.93 & 3.29 & 4.27&	5.75&	7.08&	8.72&  5.17 \\ 
					\bottomrule
				\end{tabular}
				
			}
		}
	\end{center}
	\vspace{-5ex}
\end{table*}

In our analysis, we first observe that when adapted to a Dataset-wide format, Source-free baselines such as AC, NuclearNorm, and GradNorm show significant performance improvements. Note, however, that transitioning to a Dataset-wide variant renders them no longer source-free.

Next we notice that our method consistently outperforms both Sample-Wise and Dataset-wide variants of GradNorm across all datasets examined, proving the importance of the proposed unsupervised calibration with generative model.  Although the performance of Dataset-wide baselines increases with more data, our method remains superior in scenarios with limited data availability.

Among the Dataset-wide baselines, there is no consistent leader; FID, for example, performs best Camelyon dataset, but worst on OfficeHome dataset, as shown in Table \ref{table:dataset_wide_5per}.
In the next section, we analyse and visualize different aspects that affect the performance of Dataset-wide baselines.

\subsection{Performance Analysis}\label{sec:performace_analysis}

In this section, we discuss the challenges faced by the Dataset-wide methods when access to source data is limited.

		\subsubsection{Poor Score Distribution}

\begin{figure*}[!h]
	\begin{subfigure}{0.24\textwidth}
		\includegraphics[width=\linewidth]{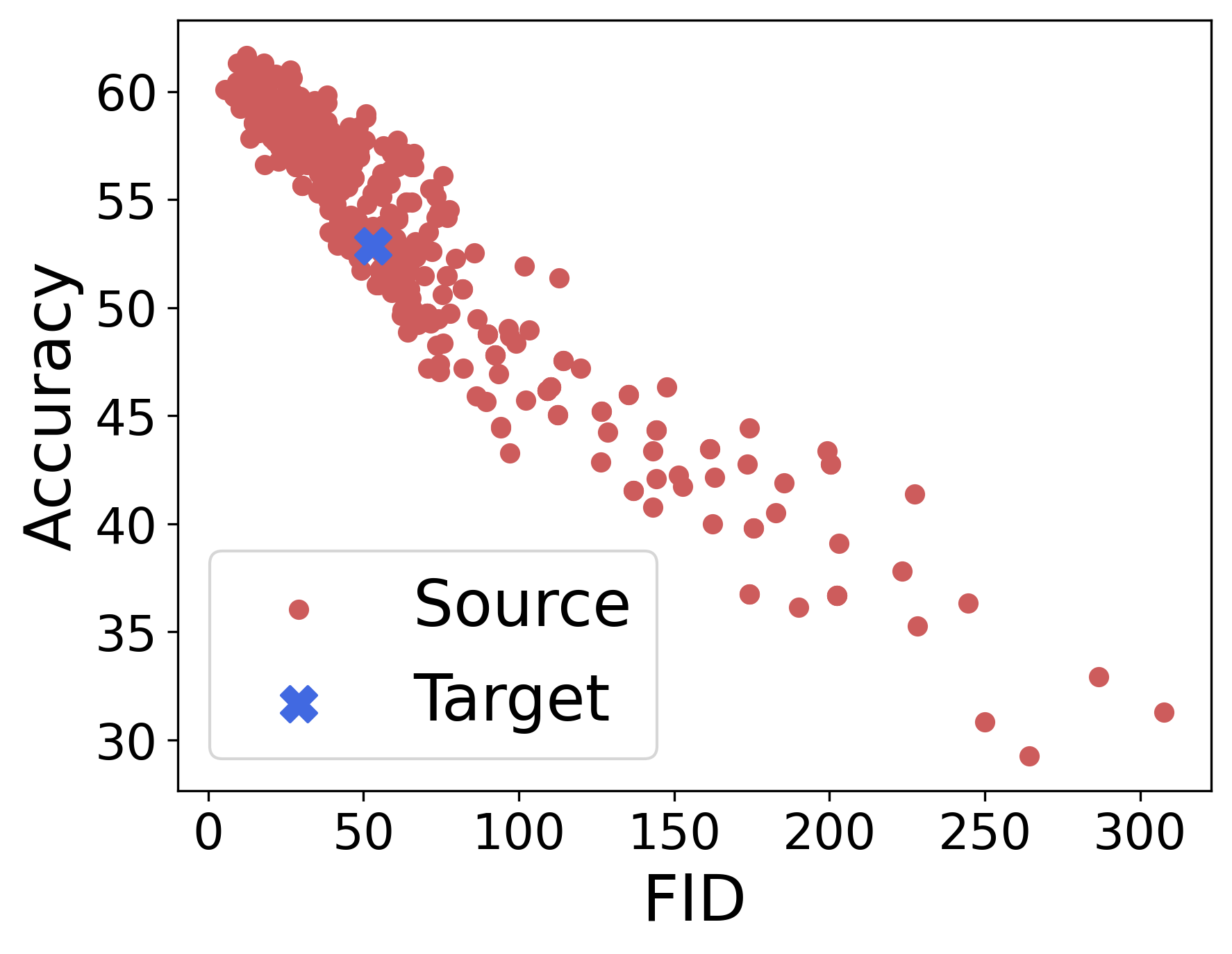}
		
		\caption{FMoW,\\Corrup. "FMoW-Test"  \\10\% openness}
		\label{img:fid_fmow_10}
	\end{subfigure}
	\begin{subfigure}{0.24\textwidth}
		\includegraphics[width=\linewidth]{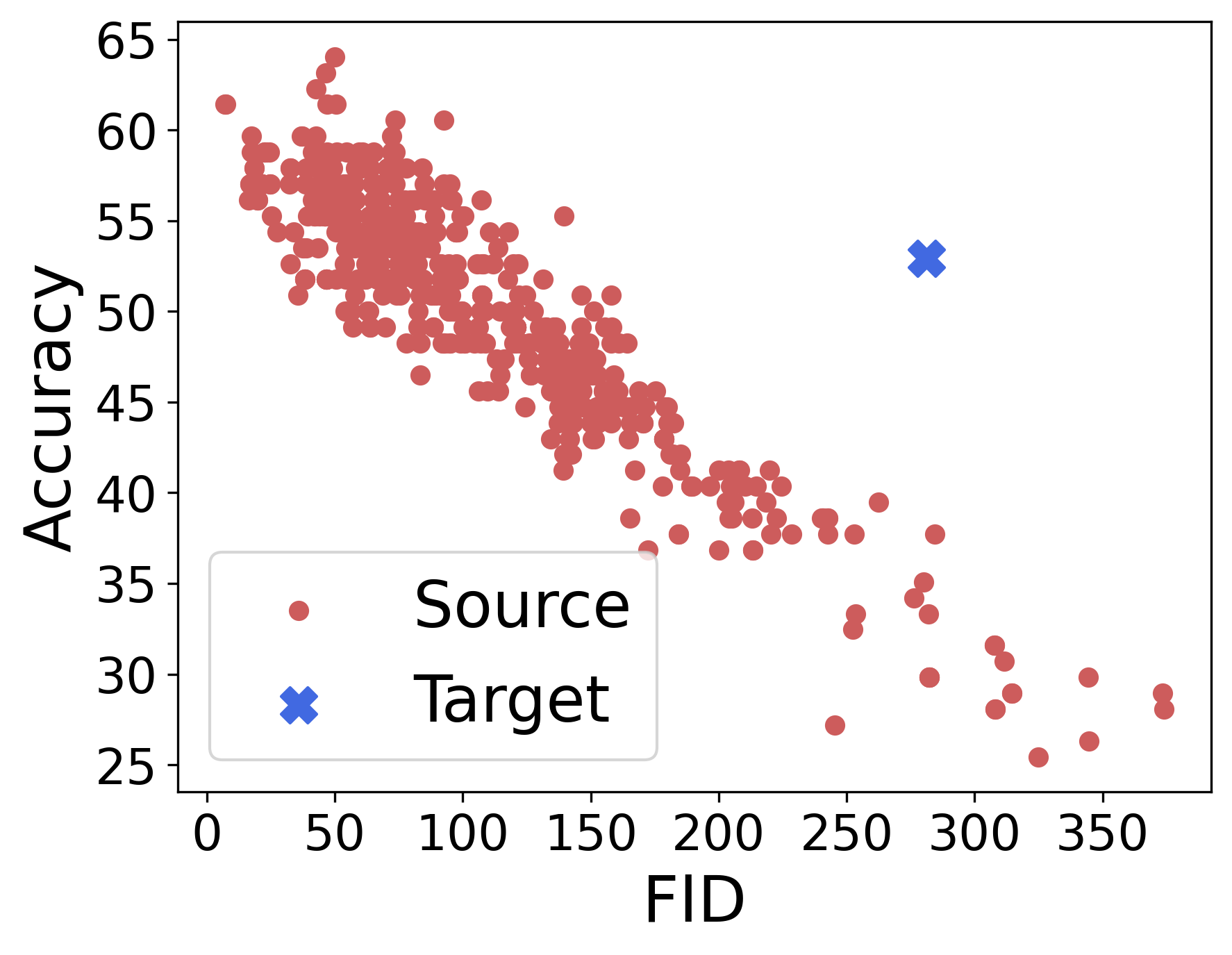}
		\caption{FMoW,\\Corrup. "FMoW-Test" \\1\% openness}
		\label{img:fid_fmow_1}
	\end{subfigure}
	\begin{subfigure}{0.24\textwidth}
		\includegraphics[width=\linewidth]{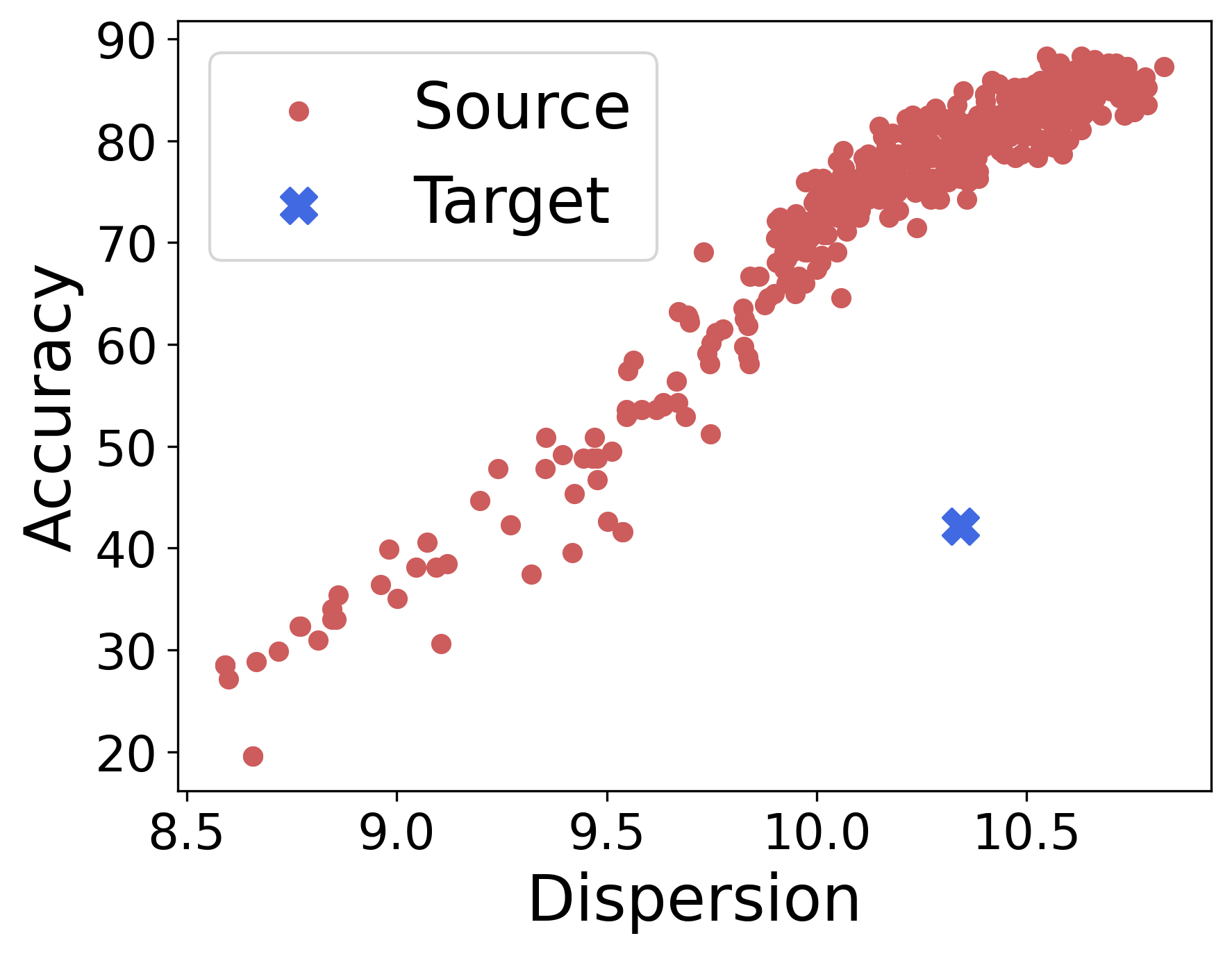}
		\caption{Terra Incognita,\\Corruption "L38" \\ 10\% openness}
		\label{img:terra_dispersion_10}
	\end{subfigure}
	\begin{subfigure}{0.24\textwidth}
		\includegraphics[width=\linewidth]{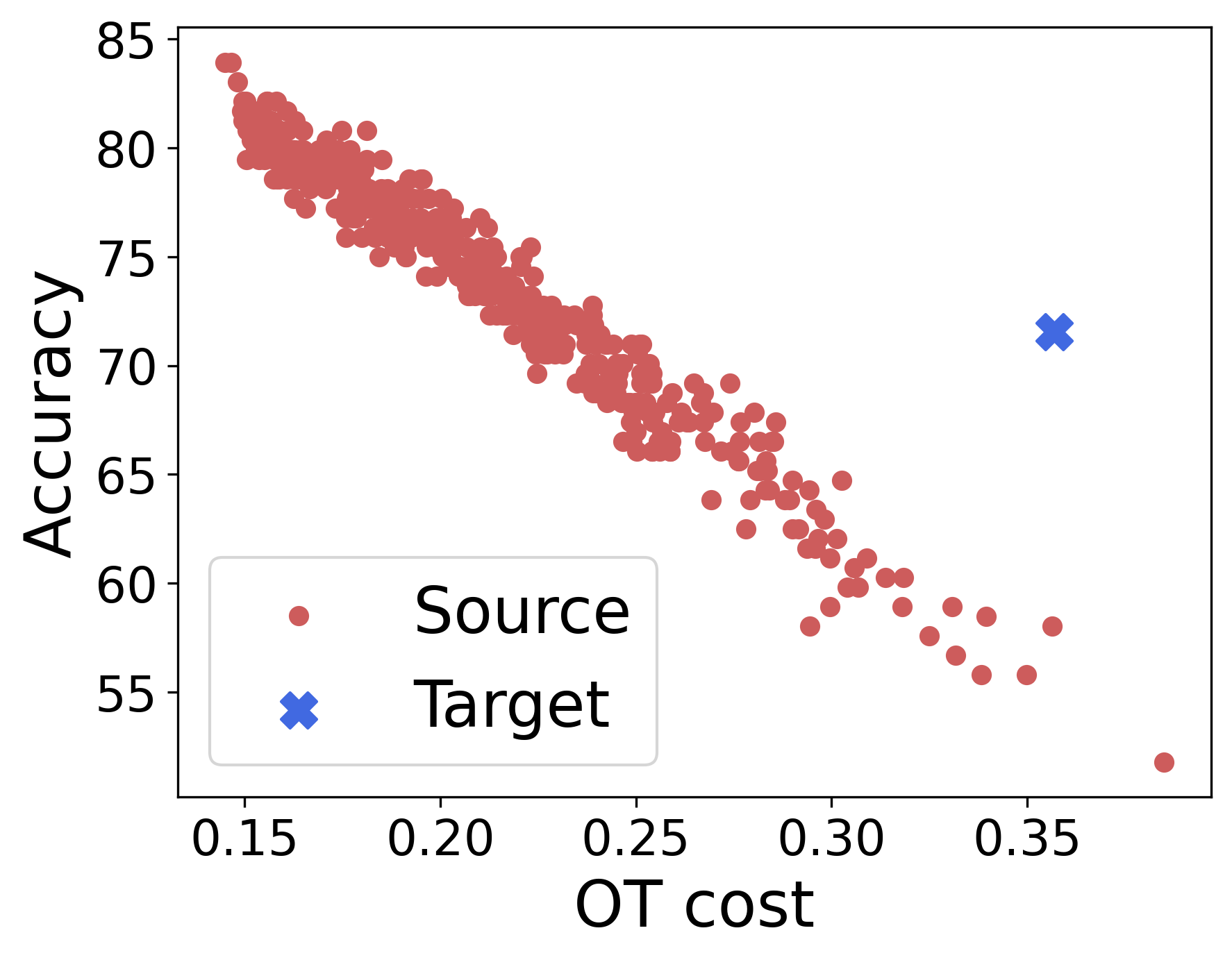}
		\caption{Office Home, \\Corruption "Photo" \\ 10\% openness}
		\label{img:office_home_ot_10}
	\end{subfigure}
	\vspace{-1ex}
	\caption{Examples of Poor Score Distribution}
	\label{img:poor_score_distr}
	  \vspace{-1ex}
\end{figure*}

We first notice that the score distributions between target and source do not always match. Moreover, the sensitivity of some scores to the sample size significantly influences their representativeness. For instance, FID, which relies on a covariance matrix, becomes less representative with a smaller number of samples. Consequently, in scenarios where openness is large, the correlation between target and source data closely aligns, as illustrated in Fig.\ref{img:fid_fmow_10}. However, when openness is reduced, the discrepancy between these distributions becomes more pronounced, negatively affecting the accuracy of the performance predictor, as seen in Fig.\ref{img:fid_fmow_1}.

\begin{figure*}[!h]
	\begin{subfigure}{0.24\textwidth}
		\includegraphics[width=\linewidth]{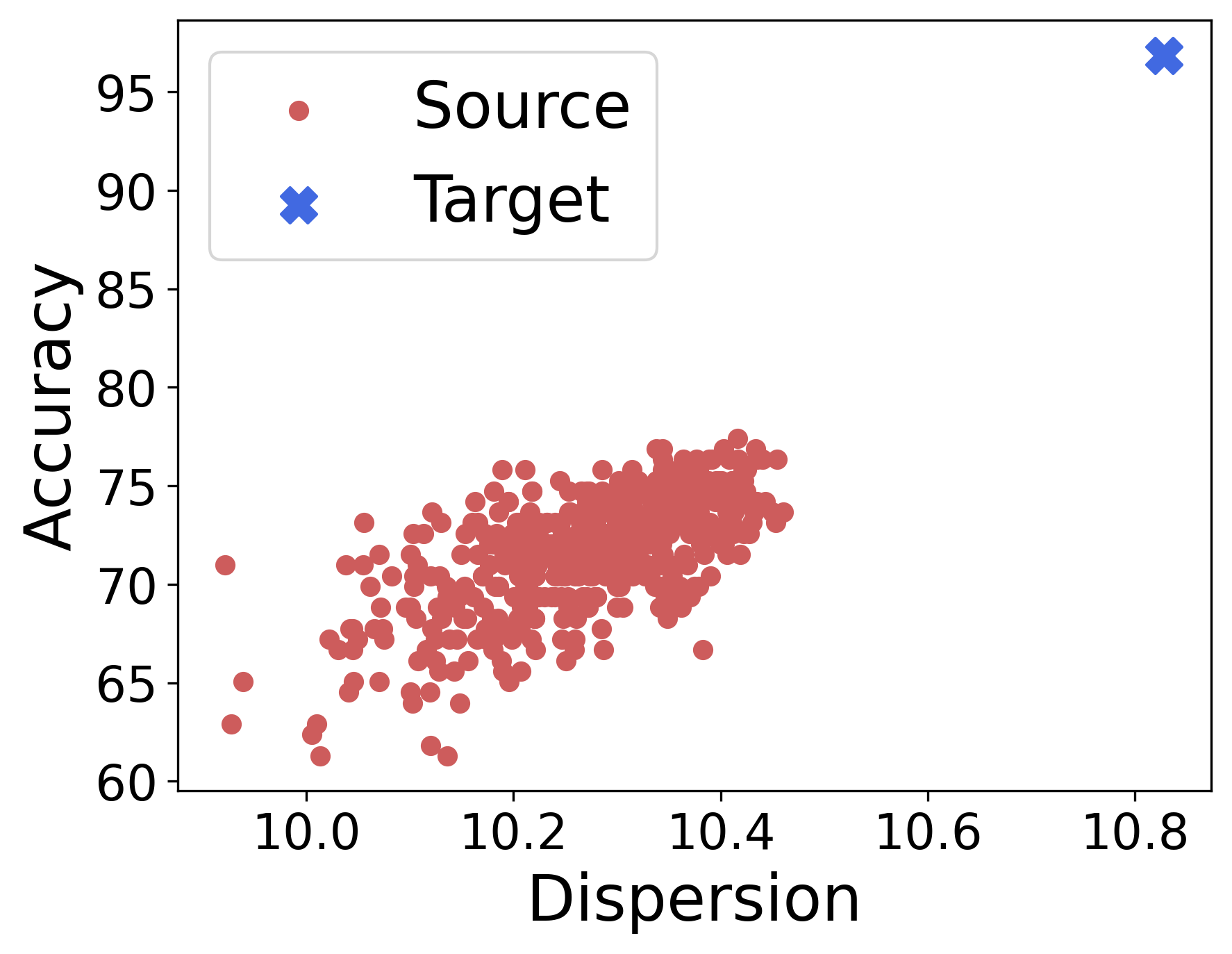}
		
		\caption{VLCS,\\Corrup. "Caltech101" \\10\% openness}
		\label{img:dispersion_vlcs_1}
	\end{subfigure}
	\begin{subfigure}{0.24\textwidth}
		\includegraphics[width=\linewidth]{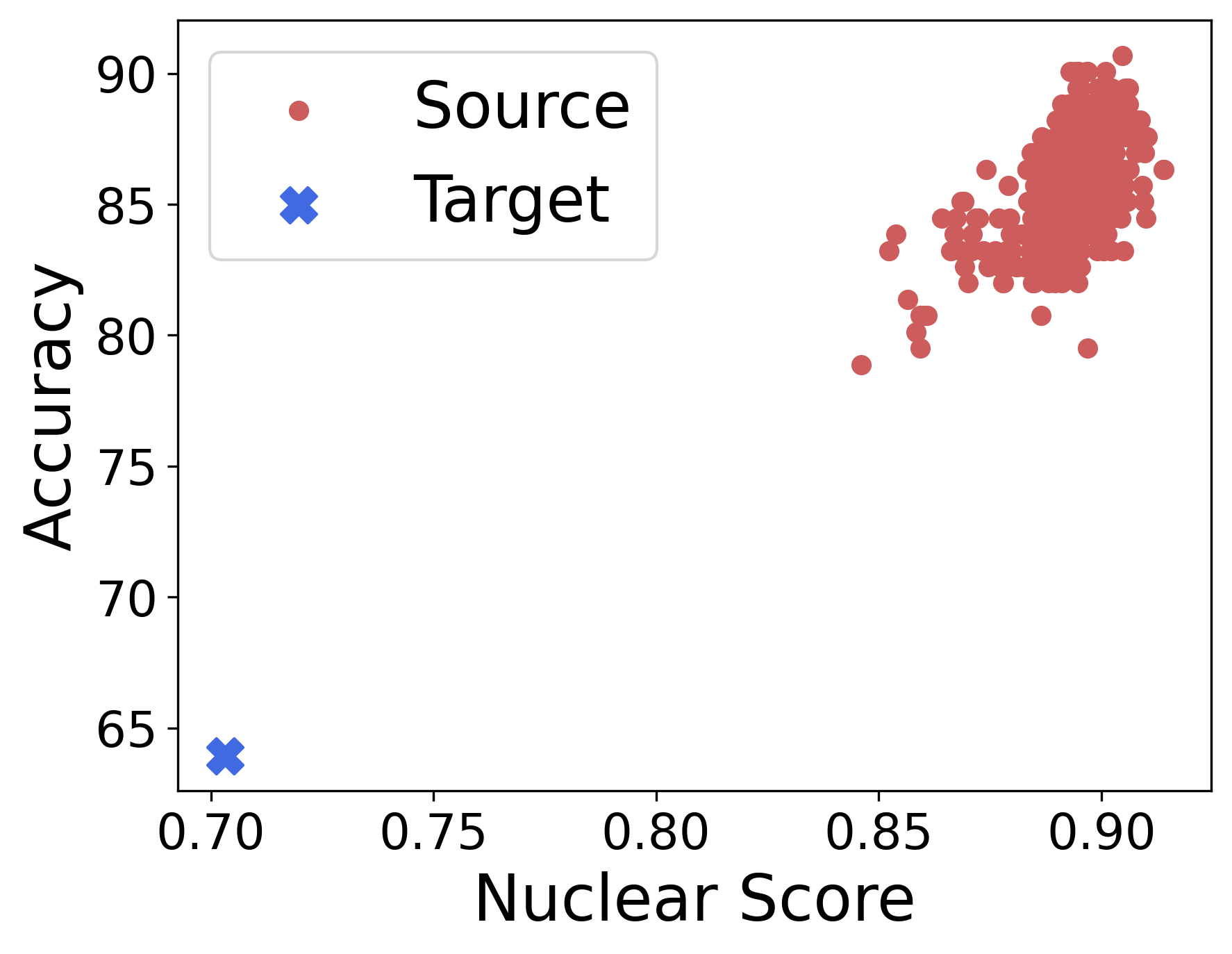}
		\caption{VLCS,\\Corrup. "LabelMe" \\10\% openness}
		\label{img:v2}
	\end{subfigure}
	\begin{subfigure}{0.24\textwidth}
		\includegraphics[width=\linewidth]{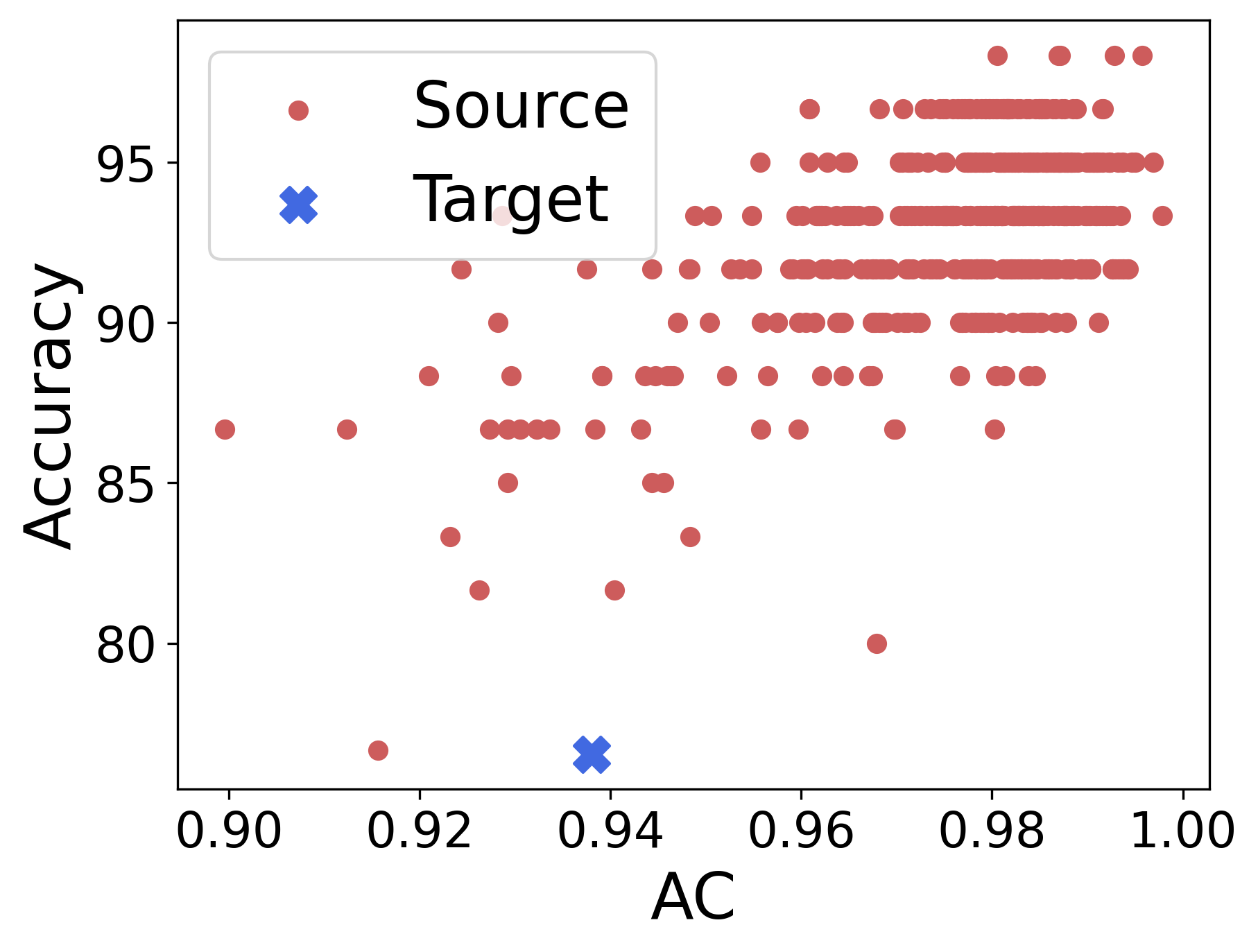}
		\caption{PACS,\\Corrup. "Photo" \\5\% openness}
		\label{img:v3}
	\end{subfigure}
	\begin{subfigure}{0.24\textwidth}
		\includegraphics[width=\linewidth]{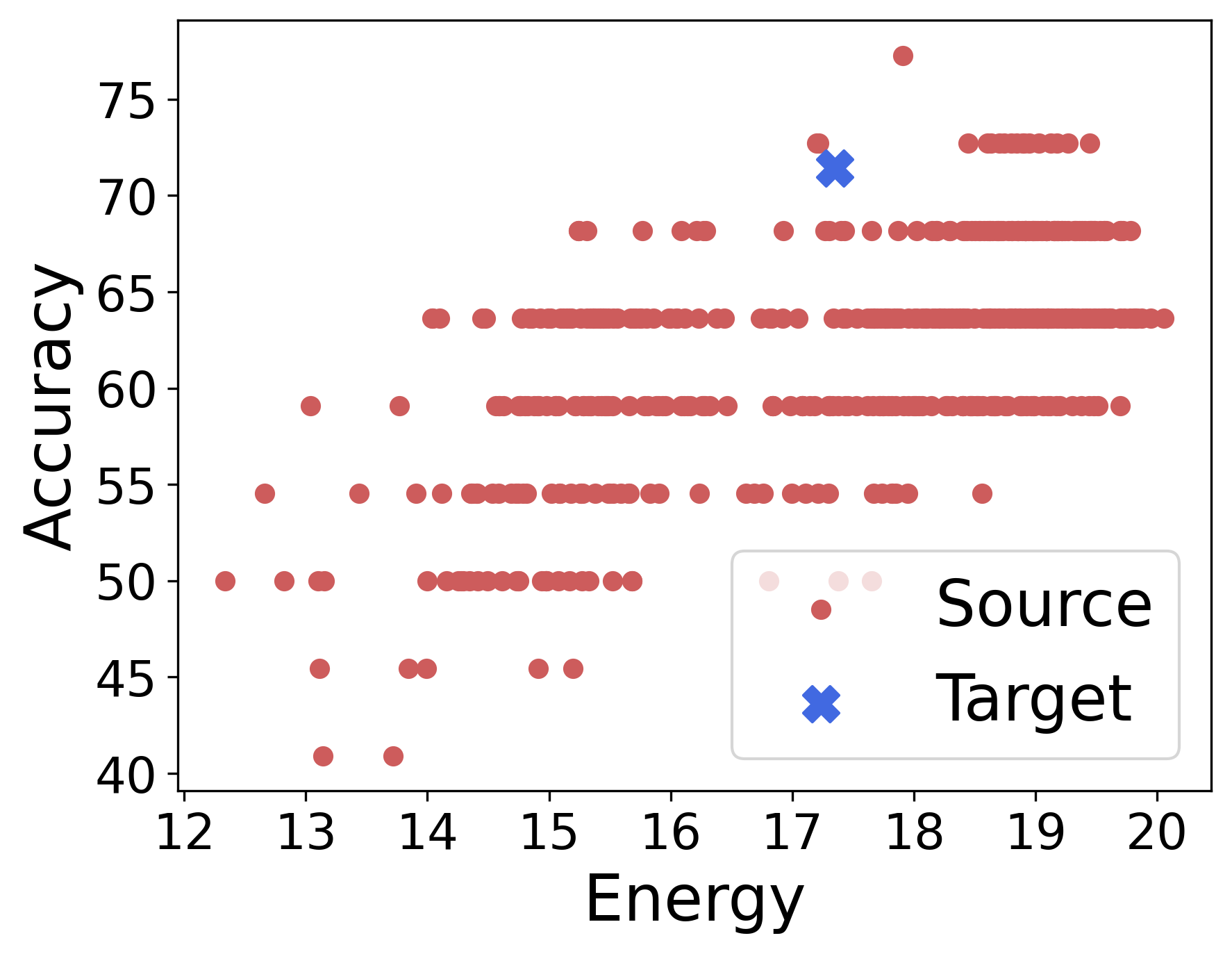}
		\caption{OfficeHome,\\Corrup. "Product" \\1\% openness}
		\label{img:v4}
	\end{subfigure}
	
	\vspace{-1ex}
	\caption{Examples of Poor Accuracy Distribution}
	\label{img:poor_acc_distr}
	\vspace{-3ex}
\end{figure*}

This discrepancy is not exclusive to scenarios with varying openness. For other datasets, even when openness is large, the score distributions between source and target significantly diverge. 
Importantly, in this work we focus on natural shifts rather than artificial shifts. We highlight that a strong correlation coefficient obtained from synthetic shifts does not necessarily translate to effective predictor performance in natural settings, as evidenced Figure~\ref{img:terra_dispersion_10} and Figure~\ref{img:office_home_ot_10}.

Given these observations, we stress the importance of utilizing Mean Absolute Error instead of solely relying on correlation coefficients for evaluating performance prediction methods. This approach provides a more reliable measure of a predictor's quality, especially in the face of natural shifts, ensuring a more accurate assessment of its effectiveness.


	\subsubsection{Poor Accuracy Distribution}

Dataset-wide methods depend on data augmentations to mimic the distribution shift between source and target data. However, when the network demonstrates robustness to these augmentations and corruptions, the accuracy of augmented source samples remains relatively unchanged. Consequently, these corrupted versions fail to accurately represent the real distribution shift. This phenomenon is depicted in Figure~\ref{img:dispersion_vlcs_1} and Figure~\ref{img:v2}, showing all source samples are clustered together far from the target samples.

Another challenge arises with a very small sample size. In such cases, the limited range of possible accuracies restricts the quality of data available for the Linear Regression model. For instance, having only 20 source samples means there are just 20 possible accuracy values. This limitation is visualized in Figure~\ref{img:v3} and  Figure~\ref{img:v4}, where the points are binned across Y axis, indicating a constrained variability in accuracy due to the small number of samples.

\end{document}